\PassOptionsToPackage{table,dvipsnames}{xcolor}
\documentclass{article} 
\usepackage{iclr2025_conference,times}

\usepackage{amsmath,amsfonts,bm}

\def\eqref#1{equation~\ref{#1}}

\def\1{\bm{1}}

\DeclareMathAlphabet{\mathsfit}{\encodingdefault}{\sfdefault}{m}{sl}
\SetMathAlphabet{\mathsfit}{bold}{\encodingdefault}{\sfdefault}{bx}{n}

\usepackage{url}
\definecolor{citecolor}{HTML}{0071bc}
\usepackage[colorlinks=true,linkcolor=red,citecolor=citecolor]{hyperref}

\usepackage[utf8]{inputenc} 
\usepackage[T1]{fontenc}    
\usepackage{booktabs}       
\usepackage{amsfonts}       
\usepackage{nicefrac}       
\usepackage{microtype}      
\usepackage{amsmath}
\usepackage{utfsym}

\usepackage{wrapfig}
\usepackage[most]{tcolorbox}
\usepackage{multirow}
\usepackage{caption}
\usepackage{tcolorbox}
\usepackage[export]{adjustbox}
\usepackage{longtable}
\usepackage{colortbl}
\usepackage{pifont}
\usepackage{algorithm}
\usepackage{algorithmic}
\usepackage{subcaption}
\usepackage{enumitem}
\usepackage{enumerate}
\usepackage{graphicx}  
\usepackage{float}

\usepackage{makecell}
\usepackage{array}
\usepackage{tabularx}
\usepackage{nameref}
\usepackage{tikz}
\usepackage{marvosym}

\definecolor{jsonblue}{RGB}{223,235,247}
\definecolor{LakeBlue}{RGB}{0,61,153}

\usetikzlibrary{shadows}
\newcommand{\circlednum}[2][red]{%
  \tikz[baseline=(char.base)]{
      \node[shape=circle, draw=none, fill=#1, minimum size=0.9em, 
        inner sep=0pt, text=white, font=\bfseries\footnotesize, 
        drop shadow={shadow xshift=0.2ex, shadow yshift=-0.2ex, opacity=0.3}] (char) {#2};
  }%
}

\lstdefinestyle{json}{
  basicstyle=\ttfamily\small,
  breaklines=true,
  backgroundcolor=\color{jsonblue!30},
  frame=none,
  aboveskip=0pt,
  belowskip=0pt,
  showstringspaces=false
}

\newcommand{\ie}{{\emph{i.e.}}}

\title{MMBench-GUI: Hierarchical Multi-Platform Evaluation Framework for GUI Agents}

\author{\textbf{
    Xuehui Wang$^{2,1*}$,
    Zhenyu Wu$^{2,1*}$,
    JingJing Xie$^{3,1*}$,
    Zichen Ding$^{1*}$,
    Bowen Yang$^{4,1*}$,
}
\\
\textbf{
    Zehao Li$^{4,1*}$,
    Zhaoyang Liu$^{5,1*}$,
    Qingyun Li$^{6,1}$,
    Xuan Dong$^{7}$,
    Zhe Chen$^{8,1}$,
    Weiyun Wang$^{9,1}$,
}
\\
\textbf{
    Xiangyu Zhao$^{2,1}$,
    Jixuan Chen$^{8,1}$,
    Haodong Duan$^{1}$,
    Tianbao Xie$^{10}$,
    Chenyu Yang$^{1}$,
}
\\
\textbf{
    Shiqian Su$^{7}$,
    Yue Yu$^{7}$,
    Yuan Huang,
    Yiqian Liu,
    Xiao Zhang,
    Yanting Zhang$^{11}$
    Xiangyu Yue$^{12}$,
}
\\
\textbf{
    Weijie Su$^{1}$,
    Xizhou Zhu$^{7}$,
    Wei Shen$^{2}$\textsuperscript{\Letter},
    Jifeng Dai$^{7}$,
    Wenhai Wang$^{12,1}$\textsuperscript{\Letter}
}
\\
\\
$^1$Shanghai AI Laboratory,
$^2$Shanghai Jiao Tong University,
$^3$Xiamen University,
\\
$^4$University of Science and Technology of China,
\\
$^5$The Hong Kong University of Science and Technology,
$^6$Harbin Institute of Technology,
\\
$^7$Tsinghua University,
$^8$Nanjing University,
$^9$Fudan University,
$^{10}$University of Hong Kong\\
$^{11}$Donghua University,
$^{12}$The Chinese University of Hong Kong
}

\iclrfinalcopy 
\begin{document}

\maketitle

\begin{abstract}

We introduce MMBench-GUI, a hierarchical benchmark for evaluating GUI automation agents across Windows, macOS, Linux, iOS, Android, and Web platforms. It comprises four levels---GUI Content Understanding, Element Grounding, Task Automation, and Task Collaboration---covering essential skills for GUI agents. In addition, we propose a novel Efficiency–Quality Area (EQA) metric to assess GUI agent execution efficiency in online automation scenarios. 
Through MMBench-GUI, we identify accurate visual grounding as a critical determinant of overall task success, emphasizing the substantial benefits of modular frameworks that integrate specialized grounding modules. Furthermore, to achieve reliable GUI automation, an agent requires strong task planning and cross-platform generalization abilities, with long-context memory, a broad action space, and long-term reasoning playing a critical role. More important, task efficiency remains a critically underexplored dimension, and all models suffer from substantial inefficiencies, with excessive redundant steps even when tasks are ultimately completed. The integration of precise localization, effective planning, and early stopping strategies is indispensable to enable truly efficient and scalable GUI automation.
Our benchmark code, evaluation data, and running environment will be publicly available at \url{https://github.com/open-compass/MMBench-GUI}.
\end{abstract}

\section{Introduction}

With the rapid advancement of Vision-Language Models (VLMs) \citep{wang2024qwen2vl,chen2024expanding,bai2025qwen25vl,zhu2025internvl3,team2025kimi,coreteam2025mimovltechnicalreport}, the capability of agents to perform complex interactions within Graphical User Interfaces (GUIs) has significantly improved \citep{wu2024copilot,cheng2024seeclick,hong2024cogagent,zheng2024seeact,gou2024navigating}. These GUI agents have shown great potential in automating complex, repetitive tasks across various domains, substantially enhancing productivity \citep{xu2024aguvis,wu2024atlas,lin2024showui,qin2025ui,yang2024aria}.

However, widely adopted evaluation benchmarks \citep{zhou2023webarena,cheng2024seeclick,xie2024osworld,li2024effects,chang2024agentboard,rawles2024androidworld,li2025screenspotpro,nayak2025ui,sun2025scienceboard,xie2025scaling} currently face several critical limitations that hinder the further development of GUI agents: (1) existing benchmarks predominantly evaluate isolated capabilities and do not comprehensively analyze the agents' overall capabilities and the relationships between multiple capabilities~\citep{deng2023mind2web,cheng2024seeclick,xie2025scaling,li2025screenspotpro}; (2) current evaluation metrics primarily emphasize task accuracy and success rate, overlooking operational efficiency~\citep{zhou2023webarena,xu2024androidlab,xie2024osworld,bonatti2024windows}; (3) insufficient coverage of evaluation scenarios fails to fully represent widely used GUI systems in real-world applications~\citep{he2024webvoyager,xie2024osworld,rawles2024androidworld,sun2025scienceboard}.

\begin{figure*}[!t]
    \centering
    \includegraphics[width=\linewidth]{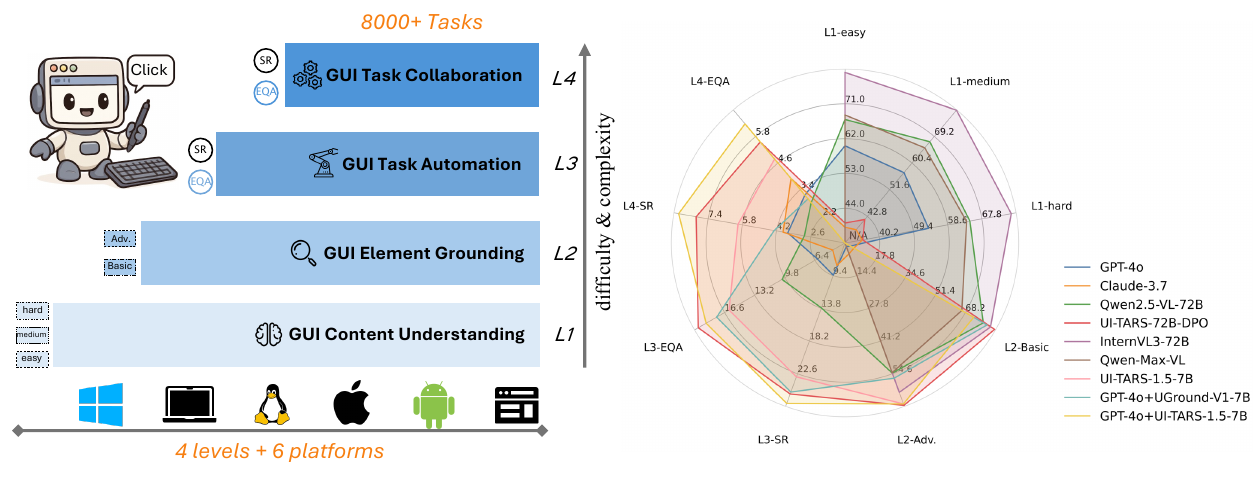}
    \caption{\textbf{MMBench-GUI}: a hierarchical benchmark spanning four levels of increasing difficulty, covering over 8,000 tasks across six commonly used platforms. From L1 to L4, task complexity increases progressively, placing growing demands on the agent’s generalization and reasoning abilities. Based on this benchmark, we visualize the performance of various models in the right figure, clearly illustrating their respective strengths as well as areas with substantial room for improvement.}
    \label{fig:MMBench-GUI}
\end{figure*}

To address these issues, we propose MMBench-GUI, a hierarchical, multi-platform benchmark framework designed for the systematic evaluation of GUI agents. As shown in Figure~\ref{fig:MMBench-GUI}, this framework comprises four progressive evaluation levels: (1) GUI Content Understanding, (2) GUI Element Grounding, (3) GUI Task Automation, and (4) GUI Task Collaboration. Each level addresses critical capabilities from basic interface understanding to complex cross-application task execution, ensuring comprehensive and systematic evaluation. Additionally, we introduce the Efficiency-Quality-Aware (EQA) metric, which evaluates both task accuracy and operational efficiency, encouraging agents to complete tasks with minimal interaction steps. Furthermore, to ensure practical relevance, we have constructed a multi-platform dataset covering Windows, macOS, Linux, iOS, Android, and Web platforms, effectively reflecting diverse real-world scenarios.

Leveraging extensive evaluations and analysis with MMBench-GUI, we identify key limitations and reveal the current state of GUI agents. Our findings indicate that: (1) although general language models excel at high-level planning and reasoning tasks, they significantly lag in precise visual interaction capabilities. Precise visual grounding is identified as a core determinant of task success, underscoring the critical need for enhanced localization accuracy; (2) efficiency now eclipses raw success rate as the next hurdle. Our EQA metric exposes the considerable surplus steps incurred by contemporary agents during task execution--redundancies that stem from localization inaccuracies, incomplete action spaces, and short-sighted or inadequate planning; (3) agent performance notably degrades when faced with complex, ambiguous, and cross-application tasks, exposing weaknesses in memory management, state tracking, and adaptive reasoning mechanisms. Addressing these shortcomings is crucial for future advancements in GUI agents. 

In summary, our primary contributions are as follows:
\begin{itemize}[leftmargin=2em]
    \item We propose a cross-platform, hierarchical benchmark designed to comprehensively evaluate GUI agents across multiple task types and difficulty levels. Inspired by a human-centered perspective, the benchmark is structured in a progressive manner, covering four essential capabilities. For static tasks (L1 and L2), we introduce fine-grained difficulty stratification to enable stepwise assessment. For dynamic tasks (L3 and L4), we provide both cleaned data splits and novel task constructions to better reflect real-world variability.
    \item We develop the first evaluation benchmark that spans all widely used operating systems, including Windows, Linux, macOS, Android, iOS, and the Web. To the best of our knowledge, this is the first work to enable consistent multi-platform evaluation of GUI agents under a unified protocol. This broad coverage allows for more realistic and application-aligned performance assessment. Notably, our benchmark is also the first to include online task scenarios for macOS, filling a long-standing gap in GUI agent evaluation.
    \item We introduce a novel metric, Efficiency-Quality-Aware (EQA), to jointly assess both the success and efficiency of agent behavior in online tasks. While most prior works focus solely on success rate (SR), EQA additionally considers when the task is completed within the step budget, providing a measure of action redundancy. This metric offers deeper insights into agent behavior and promotes the development of agents that are not only capable but also efficient—an often overlooked dimension in prior benchmarks.
\end{itemize}

\section{Related works}
\label{related_work}

\textbf{GUI Agents.}
GUI agents have attracted growing interest, driven by advances like Anthropic’s Computer-Use Agent\footnote{https://www.anthropic.com/news/3-5-models-and-computer-use} and OpenAI’s Operator\footnote{https://openai.com/index/computer-using-agent}.
Currently, GUI Agents mainly fall into two paradigms: a) Modular agent schemes~\citep{cheng2024seeclick,gou2024navigating,yang2024aria, zhang2025guimid,wu2025gui,xie2025scaling,wang2025computer}, which typically employs general-purpose VLMs (\ie, GPT-4o) as planners, integrated with a specially trained GUI grounding model for focused UI element localization; b) Native agent schemes~\citep{xu2024aguvis,wu2024atlas, lin2024showui,sun2024osgenesis,qin2025ui,yang2024aria}, where planning and grounding are trained in an end-to-end manner.
Modular approaches benefit from state-of-the-art components but face challenges in system-level alignment~\cite{cheng2024seeclick,gou2024navigating}. In contrast, the native agent paradigm aligns capabilities more naturally during training~\citep{wu2024atlas,xu2024aguvis,qin2025ui}. 
Both paradigms can use screenshots~\citep{niu2024screenagent,liu2024autoglm}, accessibility trees~(A11y Trees)~\citep{gao2023assistgui}, and HTML pages~\citep{furuta2023multimodal,deng2023mindweb} as input. 
However, A11y Trees and HTML codes vary across platforms, are prone to noise, and may cause excessive token length~\citep{zheng2024seeact, hong2024cogagent, cheng2024seeclick}.
Generally, in this work, we focus exclusively on the screenshot-only setting and propose a hierarchical, multi-platform benchmark to evaluate these vision-only native agents.

\textbf{GUI Benchmarks.}
Effectively GUIs requires a sophisticated grasp of intertwined visual and textual cues, yet this complex domain remains largely outside the scope of general-purpose multimodal QA benchmarks~\citep{liu2024mmbench,yue2024mmmu,masry2022chartqa}.
While ScreenQA~\citep{hsiao2022screenqa} and WebSRC~\citep{chen2021websrc} provide large-scale QA datasets based on Android screenshots and web pages respectively, and GUI-World introduces cross-platform GUI QA via video data, these efforts offer limited support for interactive GUI agents.
To evaluate visual grounding in GUI contexts, several benchmarks have emerged. ScreenSpot~\citep{cheng2024seeclick} and its improved versions~\citep{wu2024atlas,li2025screenspotpro} support cross-platform UI grounding with progressively enhanced realism and annotation quality. UI-I2E-Bench~\citep{liu2025ui} and UI-Vision~\citep{nayak2025ui} further expand this by aligning natural language instructions with GUI elements of varying scale and type.
For reasoning and planning, offline benchmarks like ~\citep{rawles2023androidinthewild,chen2024guicourse,li2024effects,deng2023mind2web,kapoor2024omniact,lu2024guiodyssey} assess action prediction from fixed trajectories, while online benchmarks ~\citep{zhou2023webarena,xie2024osworld,bonatti2024windows,rawles2024androidworld,xu2024androidlab,liu2024visualagentbench} enable interactive evaluation across platforms. However, macOS remains underexplored.
Our MMBench-GUI benchmark addresses this gap by enabling online evaluation on macOS and emphasizing cross-platform robustness, providing a realistic and comprehensive evaluation for GUI agents.

\section{MMBench-GUI}
\label{sec:mmbench-gui}

In this paper, we introduce MMBench-GUI, a benchmark designed to comprehensively evaluate the capabilities of AI agents in operating graphical user interface (GUI) across a broad spectrum of platforms, including Windows, Linux, macOS, Web, Android, and iOS.
Informed by a cognitive analysis of essential human abilities for GUI tasks, MMBench-GUI's evaluation process is organized into a multi-level hierarchy. The underlying principles of this hierarchical framework are discussed in Section~\ref{sec:hier_eval}.
Sections~\ref{sec:level_1}--\ref{sec:level_4} elaborate each level of MMBench-GUI, including task formulation, data sources, and evaluation protocols. %
Finally, we offer a statistical analysis of the proposed MMBench-GUI in Section~\ref{sec:statistics}.

\begin{figure*}[!t]
    \centering
    \includegraphics[width=\linewidth]{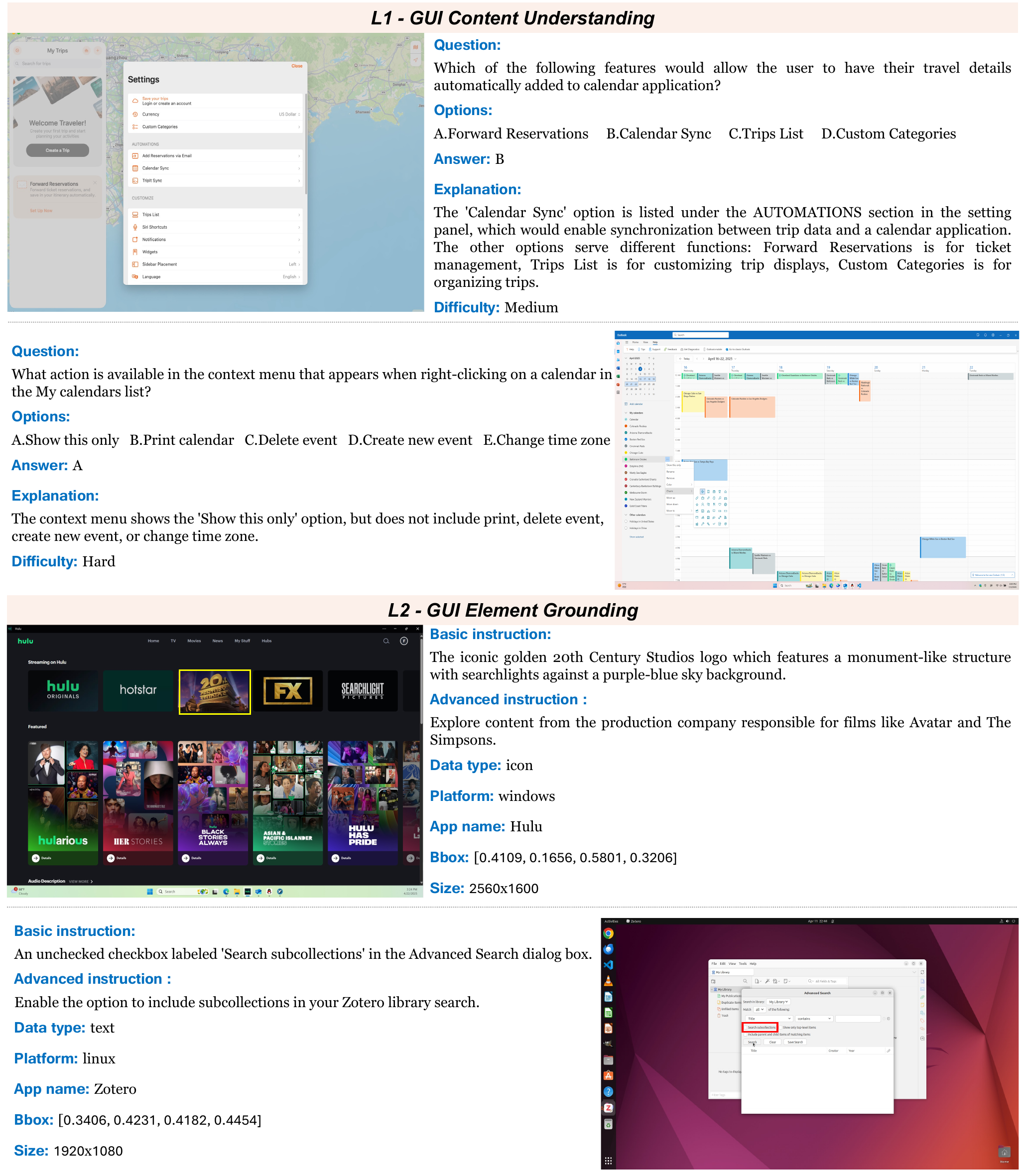}
    \caption{\textbf{Examples for L1\&L2.} Both of them are offline tasks. We provide examples from different platforms for each level. For clarity, some less critical fields are not shown here and full examples are available for download in our public repository.}
    \label{fig:level12_example}
\end{figure*}
\subsection{Hierarchical Evaluation}
\label{sec:hier_eval}
Existing GUI agents typically complete assigned tasks by emulating human operations, such as mouse clicks and keyboard input. This requires them to understand the graphical user interface and possess long-horizon planning capabilities.
However, existing benchmarks tend to focus on isolated aspects. For instance, Screenspot~\citep{cheng2024seeclick} evaluates spatial localization, while OSWorld~\citep{xie2024osworld} emphasizes end-task success—without directly assessing the full range of underlying competencies. 
As a result, the relationships among different abilities remain unclear and it is difficult to determine which specific factors contribute to an agent's success or failure. 
Taking into account these limitations and motivated by the use of leveled definitions in the domain of autonomous driving, we developed a hierarchical evaluation framework, MMBench-GUI, to systematically and comprehensively assess the capabilities of GUI Agents, as shown in Figure~\ref{fig:MMBench-GUI}.
Specifically, we organize the evaluation framework into four ascending levels: \circlednum[blue]{1} L1-GUI Content Understanding, \circlednum[blue]{2} L2-GUI Element Grounding, \circlednum[blue]{3} L3-GUI Task Automation, \circlednum[blue]{4} L4-GUI Task Collaboration.

Each level is associated with a set of tasks of increasing complexity, designed to test the Agent’s proficiency in progressively more demanding scenarios.
The complete benchmark includes over 8,000 tasks that span diverse platforms, with detailed statistics provided in Section~\ref{sec:statistics}.

\subsection{L1-GUI Content Understanding}
\label{sec:level_1}

To accurately complete an automated task based on the provided instructions, GUI agents need to integrate their domain knowledge with visual observations, enabling them to interpret the layout, functionalities, and informational content embedded in the interface. 
To achieve this goal, agents are capable of handling many challenges, such as substantial variability in UI design paradigms across different platforms (e.g., desktop \textbf{vs.} mobile), discrepancies in interface conventions among various applications even within the same platform, and fragmented background knowledge for some domain-specific software tools. 
These complexities underline the critical importance of advanced perception and understanding mechanisms. However, due to a lack of comprehensive and well-defined benchmarks, the understanding capabilities of GUI agents have not been effectively and explicitly evaluated. 

Therefore, we propose the first level of task: \textit{L1-GUI Content Understanding}. 
This task is placed at the beginning because we believe that understanding the GUI is a fundamental prerequisite to successfully complete any subsequent tasks.

\textbf{Task Definition.} At this level of task, our objective is specifically to assess the ability of an agent to extract, comprehend, and reason about information present in GUI screenshots, without requiring explicit attention to precise element localization or specific operational actions.
Therefore, we formalize this assessment as a Multiple-Choice Question-and-Answer (MCQA) task based on visual observations (GUI screenshots), enabling quantifiable and standardized output that simplifies the evaluation process.
Formally, the task can be defined as

\begin{equation}
    o^* = \text{Agent}(\mathbf{V}, q, \mathcal{O})
\end{equation}

where $\mathbf{V}$ denotes the visual observation (GUI screenshot) presented to the agent, $q$ represents the question about the observation to evaluate comprehension, and $\mathcal{O} = \{o_1, o_2, \dots, o_k\}$ represents the set of $k$ candidate options for question $q$, among which only one $o$ can correctly answer $q$.

The agent's goal is to analyze the question $q$, identify relevant information within $\mathbf{V}$, conduct reasoning, and finally select an option $o^*$ from $\mathcal{O}$ as the predicted answer.
The key to achieving this goal is the proper construction of the pair $(q,\mathcal{O})$, as the effectiveness of the evaluation depends on both the quality of the question and the relevance of the options.
Therefore, we argue that a diverse and sufficiently large set of well-constructed $(q,\mathcal{O})$ pairs about GUI elements and operations, with varying levels of difficulty, is essential to effectively evaluate the GUI understanding capabilities of the agent.

\textbf{Data Collection and Annotation.}
We manually collected screenshots from widely used applications and websites across all supported platforms, selected for their high usage frequency and representative user scenarios. 
In addition, we supplemented our data with a small number of screenshots sourced from publicly available datasets~\citep{cheng2024seeclick, li2025screenspotpro}.
To ensure diversity, we include screenshots of varying sizes, ranging from single-window to full-screen views, and accompanied each image with metadata; filenames were anonymized using an MD5-based encoding scheme constructed from a combination of platform, application name, and original file path, to avoid path conflicts and information leakage.

\begin{table*}[!t]\centering
\captionsetup{labelformat=empty}
\small
\vspace{-2mm}
\begin{minipage}{0.99\columnwidth}
    \caption{}
    \label{tab:eval_prompt}
    \vspace{-1mm}
    \centering
    \begin{tcolorbox} 
        \raggedright
        \small
        \hspace{-6mm}
    
        \textbf{Prompt 1:} \\
        \vspace{-1mm}
        \rule{\linewidth}{0.2mm} \\
        \vspace{2mm}
        
        You are an expert GUI analyst for \{os\_name\} and item-writer. \\ \vspace{2mm}
        \textbf{Input:} \\ \vspace{1mm}
        1. One screenshot of a GUI application. \\
        2. The application's name (\{app\_name\} or ``Not available") — optional and for background only; do \textbf{not} mention it in any question text. \\

        \vspace{2mm}
        \textbf{Task:} \\ \vspace{1mm}
        Create \emph{exactly one} multiple-choice question about the screenshot at \textbf{each} of three difficulty levels (easy, medium, hard).  
        For \textbf{every} question you generate: \\
        \begin{itemize}[leftmargin=2em]
            \item Write the stem in clear English that can be answered \emph{only} by understanding the screenshot. Avoid trivial facts (e.g., “What color is the button?”) unless color is functionally meaningful. 
            \vspace{-1mm}
            \item Focus on tasks, labels, hierarchy, states, or affordances shown in the UI.
            \vspace{-1mm}
            \item Provide \textbf{4--6} answer options labeled ``A'', ``B'', ``C'', … in a JSON sub-object called \texttt{"options"}.  
            \vspace{-1mm}
            \item Ensure \textbf{one and only one} option is strictly correct; the others must be clearly incorrect but plausible. Give the answer key (the letter of the correct option).  
            \vspace{-1mm}
            \item Double-check yourself that the correct answer is indeed unique and unambiguous. 
            \vspace{-1mm}
            \item Do \textbf{not} include the \{app\_name\} or any other identifying text of the app in the stem or options.
            \vspace{-1mm}
            \item Give a concise \texttt{"explanation"} stating \emph{why} the correct option is right and the others are not in 1--3 sentences. 
            \vspace{-1mm}
            \item The hard question should require the answerer to think more about the screenshot, the question, and the options (you can also make options be easy to confuse).
        \end{itemize}

        \vspace{1mm}
        \textbf{Output format:} \\ \vspace{1mm}
        
        Return a single valid JSON array containing three objects (one per difficulty), in \emph{English}, structured exactly like this schema: \\
        
    \begin{lstlisting}[style=json]
[
  {
    "difficulty": "easy",
    "question": "<stem>",
    "options": {
      "A": "<option text>",
      "B": "<option text>",
      "C": "<option text>",
      "D": "<option text>"   // add "E","F" only if needed
    },
    "answer": "A",
    "explanation": "<brief rationale>"
  },
  ...
]
    \end{lstlisting}

    \vspace{2mm}
        \textbf{Important Constraints:} \\ \vspace{1mm}
        
        1. Produce only the JSON text—no markdown, headings, or commentary. \\
        2. Validate that the JSON is syntactically correct before outputting. \\
        3. After generation, internally review each Q\&A for accuracy and compliance.

    \end{tcolorbox}
\end{minipage}
\end{table*}

Then, we followed a four-step strategy to construct high-quality Question-Options-Answer pairs:
\begin{itemize}
    \item Step 1: Claude 3.7~\citep{anthropic2025claude37} was used to generate three questions for each image, corresponding to three levels of difficulty: easy, medium, and hard. Each question includes 4 to 6 answer options, with exactly one correct choice. In addition, Claude 3.7 was instructed to provide an explanation for each question, detailing the reasoning process that leads to the correct answer. In designing the questions for each image, we guided Claude 3.7 to focus on various aspects of the GUI, including the functionality of UI elements, structural relationships within the interface, content states, hierarchical layout, and executable tasks. The detailed prompt for this process is shown in Prompt~\ref{tab:eval_prompt}.
    \item Step 2: We then used GPT-o4-mini~\citep{openai2025introducing} to verify the validity of each question, set of options, and answer, jointly considering the UI interface and the generated explanation. The errors were corrected with justification and revised explanations.
    \item Step 3: Then, GPT-o3~\citep{openai2025introducing} was used to further review and refine the revised items following the same Prompt~\ref{tab:eval_prompt2} as in Step 2.
    \item Step 4: Finally, manual sampling was performed to ensure overall quality and consistency.
\end{itemize}

\begin{table*}[!b]\centering
\captionsetup{labelformat=empty}
\small
\vspace{-2mm}
\begin{minipage}{0.99\columnwidth}
    \caption{}
    \label{tab:eval_prompt2}
    \vspace{-1mm}
    \centering
    \begin{tcolorbox} 
        \raggedright
        \small
        \hspace{-6mm}
    
        \textbf{Prompt 2:} \\
        \vspace{-1mm}
        \rule{\linewidth}{0.2mm} \\
        \vspace{2mm}
        
        You are a meticulous GUI-QA evaluator. \\ \vspace{2mm}
        \textbf{Input:} \\ \vspace{1mm}
        1. One screenshot (\texttt{image}) of a GUI application running on a \{os\_name\}. \\
        2. The application’s name (\texttt{app\_name}) – optional and strictly for background; \emph{never} mention it in your output. \\
        3. A JSON-like array (\texttt{qa\_items}) containing three single-choice questions about the screenshot (intended levels: \texttt{easy}, \texttt{medium}, \texttt{hard}).  
        Each object is expected to have the keys question, options, answer, difficulty, and optionally explanation. \\ \vspace{1mm}
        * Ignore cosmetic or syntactic issues in the supplied JSON (e.g., extra backticks, missing quotes, inconsistent key order, markdown fences).  \\
        * Focus \textbf{only} on the content of \texttt{question}, \texttt{options}, and \texttt{answer} when deciding validity.\\

        \vspace{2mm}
        \textbf{Task:} \\ \vspace{1mm}
        For each question, decide whether it is content-valid for use in a test. A question is valid only if all the following hold: \\
        \begin{itemize}[leftmargin=2em]
        \item The stem can be answered solely by inspecting the screenshot (no outside knowledge).
        \vspace{-1mm}
        \item Exactly one option is correct and that option is the one listed in \texttt{answer}.
        \vspace{-1mm}
        \item Incorrect options are clearly wrong yet still plausible.
        \vspace{-1mm}
        \item Neither stem nor options reveal the \texttt{app\_name}.
        \vspace{-1mm}
        \item The difficulty label is reasonable (honor system; do not reject only for minor mislabelling).
        \end{itemize}
        The hard level should allow the answerer to think more deeply about the screenshot, the question, and the options. You may make the options easy to confuse. \\
        * Do \textbf{not penalise} minor formatting faults that do not affect the five substantive criteria. \\ \vspace{2mm}
        
        \textbf{Output format:} \\ \vspace{1mm}
        
        Return a JSON array of three objects in the original order, each with: \\
        
    \begin{lstlisting}[style=json]
{
  "difficulty": "<same as input>",
  "valid": "yes" | "no",
  "comment": "<if valid: empty string; if not valid: brief reason why>",
  "fix": <if valid: null; if not valid: a *fully corrected* object that replaces the faulty one (same schema as above, with all issues fixed)>
}
    \end{lstlisting}

    \vspace{2mm}
        \textbf{Notes:} \\ \vspace{1mm}
        
        1. Provide an empty string (`` ") for \texttt{comment} and ``null" for \texttt{fix} when \texttt{valid} is ``yes". \\
        
        2. When \texttt{valid} is ``no", supply both an actionable \texttt{comment} and a complete \texttt{fix} object that meets all criteria. \\
        
        3. Do not wrap the result in markdown or add explanations outside the JSON. \\

        4. Verify that the final JSON is syntactically correct before sending it.

    \end{tcolorbox}
\end{minipage}
\end{table*}

By incorporating three different strong models across the pipeline, we reduced the risk of model-specific hallucinations and stylistic bias. 
We provide some example in Figure~\ref{fig:level12_example}, which contains screenshots, metadata, and generated annotations.

\textbf{Evaluation Metrics.} 
For each question, we adopt accuracy as the evaluation metric, consistent with common QA tasks. 
Formally, the accuracy for an evaluation set comprising $N$ Question-Options-Answer pairs can be defined as:
\begin{equation}
    \text{Acc} = \frac{1}{N} \sum_{i=1}^{N} \Theta(o^*_i = o_i),
\label{Eq2}
\end{equation}

where $\Theta(o^*_i = o_i)$ is an indicator function that equals 1 if the predicted answer $o^*_i$ for the $i$-th pair matches the ground-truth answer $o_i$ and 0 otherwise.

To account for variations in the number of answer choices, we introduce a simple dynamic adjustment factor $\alpha$ to rescale the original accuracy of each question. Taking Windows platform which has $N_{win}$ questions as an example, the accuracy of L1 is computed as:
\begin{equation}
\text{Acc}_{win} = \frac{1}{N_{win}} \sum_{i=1}^{N_{win}} \alpha \cdot \Theta(o^*_i = o_i), \ \ \ \alpha = \frac{m_i-1}{m_i}
\end{equation}
where $m_i$ is the number of options for question $i$. Accordingly, for any given difficulty level, the agent’s understanding ability (i.e., accuracy) can be computed as:
\begin{equation}
\text{Score} = \sum_{j \in \mathcal{O}} \frac{N_j}{N} \cdot \text{Acc}_j
\label{eq:eq_4}
\end{equation}
where $\mathcal{O} = \{\texttt{win}, \texttt{linux}, \texttt{mac}, \texttt{ios}, \texttt{android}, \texttt{web}\}$ denotes the set of operation platforms, $N_j$ is the number of questions for platform $j$, $N = \sum_{j \in \mathcal{O}}{N_j}$ is the total number of questions across all platforms.

\subsection{Level 2: GUI Element Grounding}
\label{sec:level_2}

The precise grounding of interactive UI elements is a fundamental prerequisite for effective execution of GUI-based tasks. 
This capability requires agents to accurately localize the spatial positions of target elements within the GUI, conditioned on the current task objective and corresponding observation (e.g., a screenshot). 
Despite significant progress in this direction, several inherent challenges remain: (1) visual ambiguity caused by highly similar elements, such as identical buttons or icons with only subtle differences; (2) dynamic UI disruptions, including pop-up windows or transient notifications that obscure intended targets; and (3) the difficulty of distinguishing inactive or grayed-out regions from their active counterparts.
Addressing these issues is crucial, as grounding directly influences an agent's reliability and effectiveness in performing GUI-based tasks. 

However, existing benchmarks such as ScreenSpot~\citep{cheng2024seeclick,wu2024atlas} have become nearly saturated, and ScreenSpot Pro~\citep{li2025screenspotpro} is curated within limited application domains. 
More critically, the instructions employed by current benchmarks are often overly simplistic and direct (e.g. ``submit the paper"), which fails to reflect the nuanced ways in which GUI agents refer to and reason about UI elements during real-world task execution.
This mismatch results in a gap between the benchmark tasks and the genuine challenges faced by agents in practical scenarios.
To address these limitations, we draw inspiration from the strengths of prior benchmarks while introducing a new dataset encompassing a broader range of application domains and more diverse and realistic instructions.
Specifically, we systematically categorize instructions by their descriptive types, aiming to more accurately assess model weaknesses and bridge the gap between benchmark evaluation and real-world agent reasoning.

\textbf{Task Definition.} Accurate perception and understanding by an agent typically require validation through concrete actions, analogous to human interactions with GUI elements, to execute subsequent task steps. 
Building upon the comprehension capabilities assessed in L1, we propose \textit{L2-GUI Element Grounding} to further measure the agent's spatial localization ability, specifically, the accurate identification of actionable GUI elements. 
This ability aligns precisely with the requirements of a grounding task, formally defined as:

\begin{equation}
    p=\text{Agent}(\texttt{ins}, \mathbf{V})
\label{Eq3}
\end{equation}

where $\texttt{ins}$ represents an instruction for the GUI element to be localized, which can be derived from a direct user task or the agent’s internal reasoning process. 
The output $p$ denotes the resulting location of the target element, typically represented by the coordinates $(x,y)$ that indicate the activation point of the interactive element. 
The definition of $\texttt{ins}$ constitutes the core component of this level. 
In the context of GUI tasks, the description of an element can encompass various attributes, including appearance, approximate spatial position, and functionality.

\begin{table*}[!t]\centering
\captionsetup{labelformat=empty}
\small
\vspace{-2mm}
\begin{minipage}{0.99\columnwidth}
    \caption{}
    \label{tab:eval_prompt3}
    \vspace{-1mm}
    \centering
    \begin{tcolorbox} 
        \raggedright
        \small
        \hspace{-6mm}
    
        \textbf{Prompt 3:} \\
        \vspace{-1mm}
        \rule{\linewidth}{0.2mm} \\
        \vspace{2mm}
        
        You are a GUI agent currently operating on a \{os\_name\}. \\ \vspace{2mm}
        \textbf{Input:} \\ \vspace{1mm}
        1. The first image is a screenshot from the \{application\} \{app\_or\_web\}, in which a selected element is highlighted with a distinctive red box and a red arrow. \\
        2. The second image is the cropped region containing the selected element and corresponding box and arrow. \\
        3. A simple and coarse description of the selected element. \\ \vspace{1mm}

        \vspace{2mm}
        \textbf{Task:} \\ \vspace{1mm}
        Your task is to understand the possible role, function, and related global contextual information of the selected element on the current page from the first image. Then, from the second image, you can combine the global information from the first image to further analyze the relationship between the selected element and its surrounding information. The simple and coarse description can be regarded as a prior for the selected element. Finally, you are required to conclude two types of instructions for the selected element: \\  \vspace{1mm}
        \textit{* Basic Instruction}: Informative description that summarizes key information. \\
        \textit{* Advanced Instruction}: An indirect yet specific instruction that refers to the selected element. \\
        
        \vspace{2mm}
        \textbf{Guidelines for Generating Descriptions:} \\ \vspace{1mm}
        \textit{Basic Instruction}: \\
        \begin{itemize}[leftmargin=2em]
          \item Concise summary including appearance and position.
          \vspace{-1mm}
          \item Avoid referencing the red box or arrow.
          \vspace{-1mm}
          \item Examples: 
          \vspace{-1mm}
            \begin{itemize}[leftmargin=1.5em]
            \item ``A circular icon with a white background and a magnifying glass symbol in black." 
            \vspace{-1mm}
            \item ``Located in the top-right corner, to the right of the profile avatar icon." 
            \end{itemize}
        \end{itemize}
        \textit{Advanced Instruction}: \\
        \begin{itemize}[leftmargin=2em]
            \item Focus on function and reasoning.
            \vspace{-1mm}
            \item Avoid visual/positional terms.
            \vspace{-1mm}
            \item Examples:
            \vspace{-1mm}
                \begin{itemize}[leftmargin=1.5em]
                    \item "Search some latest posts"
                    \vspace{-1mm}
                    \item "Type in text to discover related content"
                \end{itemize}
        \end{itemize}
        
        \vspace{1mm}
        \textbf{Output format:} \\ \vspace{1mm}
        
        Return a dictionary with: \\
        
    \begin{lstlisting}[style=json]
{
   "basic_instruction": ["xxxx", "xxx", "xxx"],
   "advanced_instruction": ["xxxxx", "xxx", "xxx"]
}
    \end{lstlisting}

    \vspace{2mm}
        \textbf{Notes:} \\ \vspace{1mm}
        
        1. Ensure instructions are clear, unambiguous, and concise. \\
        2. Do not mention the red box and arrow. \\
        3. Coarse descriptions are only priors. 

    \end{tcolorbox}
\end{minipage}
\end{table*}

\textbf{Data Collection and Annotation.} We reuse the data from L1 to annotate additional agent capabilities, enabling multidimensional analysis on a consistent data foundation. 
This design facilitates exploration of inter-task correlations and addresses earlier research questions. 
We also manually labeled the positions of interactive elements, i.e. user-operable components such as buttons or icons, using bounding boxes, and categorized them as either Text or Icon, following the classification scheme used in ScreenSpot~\citep{cheng2024seeclick}.

We adopt a three-step procedure to generate grounding instructions for annotated interactive elements:
\begin{itemize}
    \item Step 1: Claude 3.7 was prompted to produce two types of instruction per element: \textbf{Basic}, which describes visual features and approximate location to test perception-based grounding, and \textbf{Advanced}, which targets functional understanding through implicit cues. To increase diversity, three stylistic variants were generated for each type. The detailed prompt for this step is shown in Prompt~\ref{tab:eval_prompt3}.
    \item Step 2: We developed an annotation tool to manually review and refine these instructions, ensuring that each uniquely maps to a specific element. 
    \item Step 3: A validated instruction per type was selected to form the final evaluation set. 
\end{itemize}

Examples of annotated data can be found in Figure~\ref{fig:level12_example}, and we attach two types of instructions for each element of a screenshot.

\textbf{Evaluation Metrics.} Following the evaluation protocol of ScreenSpot~\citep{cheng2024seeclick}, we computed accuracy separately for the Basic and Advanced instruction types. 
For each interactive element, a prediction was considered successful if the agent's predicted point of interaction--represented as a coordinate $(x,y)$--fell within the annotated bounding box. 
Otherwise, it was marked as a failure. The final accuracy was calculated as the proportion of successful predictions over the total number of evaluated elements.

\subsection{Level 3: GUI Task Automation}
\label{sec:level_3}

\begin{figure*}[!t]
    \centering
    \includegraphics[width=\linewidth]{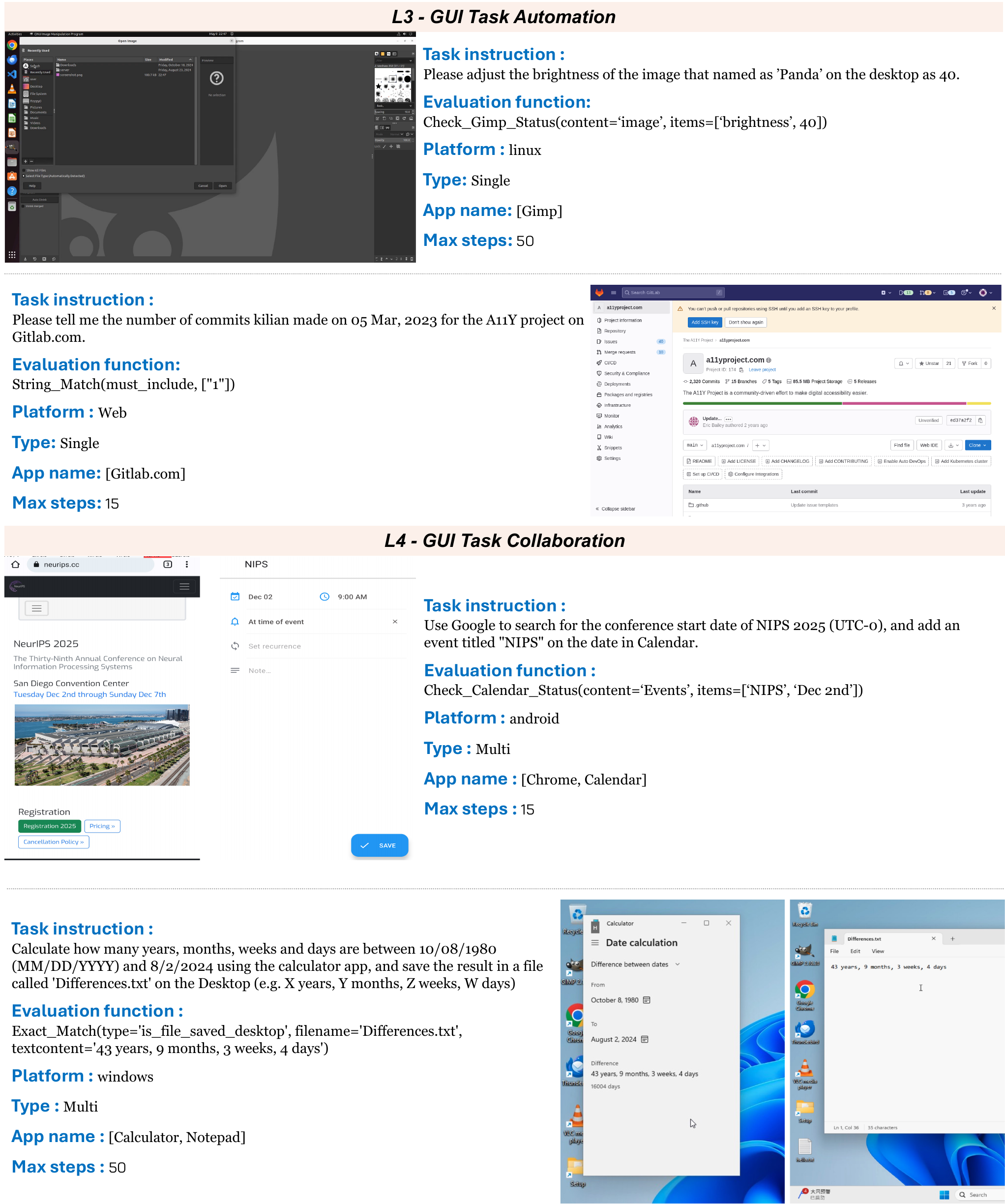}
    \caption{\textbf{Examples for L3\&L4.} Tasks of these levels are evaluated in the virtual environment with an online manner. In L4, we provide two images belonging to different applications as examples to demonstrate that collaboration is the core aspect for this level.}
    \vspace{-2mm}
    \label{fig:level34_example}
\end{figure*}

To successfully accomplish user-specified tasks within a single application environment, agents must integrate their comprehension of the interface content and precise localization of relevant elements with advanced planning and dynamic reasoning. 
The typical workflow begins with interpreting the task instruction, perceiving the content, and grounding the target UI components. 
The agent then decomposes the high-level task into a sequence of executable actions, such as clicking, typing, or selecting, iteratively interacting with the environment and adapting its strategy based on real-time feedback. This tightly coupled cycle of perception, decision-making, and interaction constitutes the essence of robust GUI task completion, especially for complex, multi-step scenarios.

Key challenges at this iterative cycle include resolving ambiguous or under-specified instructions, navigating dynamic UI states (e.g., pop-ups, context changes), and efficiently planning multi-step operations to achieve the desired outcome. Despite their prevalence in real-world automation scenarios, such capabilities within the single application are rarely evaluated in a systematic manner across multi-platforms.
Therefore, we propose \textit{L3-GUI Task Automation} as the third level in our benchmark, focusing on the agent’s ability to perform end-to-end automation within a single, potentially complex, application environment. This level serves as a crucial bridge between low-level perception/understanding and higher-level, generalizable task-solving skills.

\textbf{Task Definition.} 
Building upon the challenges outlined above, we formally define the \textit{L3-GUI Task Automation} as follows:
The agent is required to complete a multi-step task within a single application by generating a sequence of actions that directly manipulate the user interface to fulfill a specified objective. At each time step $t$, the agent receives a visual observation $\mathbf{V}_t$ of the current UI state and generates an action $A_t$ with corresponding parameters $P_t$, based on the task instruction $\texttt{ins}$, the history $\mathcal{H}_t$, and involved applications $\mathcal{S}$ (with $\mathcal{S} \in  \{\texttt{App}_1, \texttt{App}_2, \dots, \texttt{App}_n\}$ for single-app scenarios). The process is formally described as:

\begin{equation}
\begin{split}
A_t, P_t &= \text{Agent}(\texttt{ins}, \mathbf{V}_t, \mathcal{H}_t, \mathcal{S}_s) \\
\mathbf{V}_{t+1} &= \text{Env}(A_t, P_t) \\
\mathcal{H}_{t+1} &= \{\mathcal{H}_t, (V_t, A_t, P_t)\}
\end{split}
\label{Eq4}
\end{equation}

Here, $\mathcal{H}_t$ denotes the contextual history, which normally consists of previous observations and action sequences. 
In practice, the implementation of history typically follows two styles. 
The first style encapsulates the entire interaction process within multi-turn dialogues, while the second one condenses history into natural language and  injects it into the prompt.
The agent-environment interaction proceeds iteratively until a maximum number of steps ($t = T_{\max}$) is reached or a terminal action ($A_t \in$ \texttt{[FINISH, FAIL]}) is predicted.

\textbf{Task Collection and Curation.}
To ensure broad coverage and real-world relevance, our GUI task automation benchmark encompasses tasks across multiple major platforms, including Windows, Linux, macOS, web, and Android. Due to the inherent restrictions of the iOS ecosystem, iOS tasks are not currently included.

The majority of tasks are sourced from established public benchmarks, each of which leverages virtualization technology to provide robust and reproducible GUI environments. Specifically, tasks for the Linux platform are drawn from OSWorld~\citep{xie2024osworld}, Android from AndroidWorld~\citep{rawles2024androidworld}, web from WebArena~\citep{zhou2023webarena}, and Windows from WindowsAgentArena~\citep{bonatti2024windows}. These resources have been extensively validated in prior research and collectively provide a diverse set of task scenarios.
Importantly, our use of these benchmarks is not a simple replication. Each task underwent a rigorous manual review process, during which we excluded any instances likely to result in agent failure due to non-agent factors such as unstable network conditions, required account authentication, or platform-specific anomalies. This curation ensures that the performance evaluations reflect true capabilities of the agent, rather than artifacts of the benchmarking environment.

To address the lack of existing online evaluation resources for the macOS platform, we introduce \textit{MMBench-GUI-macOS}, a novel set of 70 curated tasks spanning 9 widely used macOS applications. Of these, 35 tasks are categorized as L3 tasks and the remaining 35 as L4 tasks. Task design for macOS follows the same principles as for other platforms, utilizing paired natural language instructions and screenshots to simulate virtual environments, thereby ensuring consistency and comparability across all platforms. This multi-platform, carefully curated task set provides a comprehensive and fair foundation for benchmarking GUI agents in realistic and heterogeneous settings. We provide two illustrative examples in the upper part of Figure~\ref{fig:level34_example} to demonstrate the details of L3 tasks.

\textbf{Evaluation Metrics.} From a user-centric standpoint, an ideal agent should be both accurate and efficient. However, existing benchmarks typically rely solely on Success Rate (SR), neglecting how quickly tasks are completed. To address this limitation, we propose the \textbf{Efficiency–Quality Area (EQA)}, a unified metric inspired by the AP computation protocol in COCO~\citep{lin2014microsoft}. EQA jointly considers task success and completion speed, rewarding agents that solve more tasks using fewer steps.

Specifically, we define EQA as a continuous-time recall metric over cumulative agent effort. Consider an ordered set of $N$ tasks. For each task $i \in \{1,2,\dots,N\}$, let:

\begin{itemize}
    \item $s_i = 1$ if the agent successfully completes task $i$, and $s_i = 0$ otherwise,
    \item $t_i > 0$ be the number of steps the agent takes to complete task $i$.
\end{itemize}

We define the cumulative cost and cumulative success after the first $k$ tasks as:
\begin{equation}
    T_k = \sum_{j=1}^k t_j, 
    \qquad 
    S_k = \sum_{j=1}^k s_j.
\label{Eq5}
\end{equation}

Let the global budget be $T_{\text{max}} = N \cdot t_{\text{max}}$, where $t_{\text{max}}$ is the maximum step limit per task. We normalize the cumulative effort as:
\begin{equation}
    u_k = \frac{T_k}{T_{\text{max}}} \in [0, 1].
\label{Eq6}
\end{equation}

The instantaneous recall at normalized time $u$ is defined as:
\begin{equation}
    R(u) = \max_{k:\, u_k \le u} \frac{S_k}{N}, 
    \qquad 
    u \in [0,1].
\label{Eq7}
\end{equation}

Finally, EQA is computed as the area under the step-wise non-decreasing recall curve:
\begin{equation}
    \mathrm{EQA} = \int_0^1 R(u)\,du 
    \;\approx\; 
    \frac{1}{M} \sum_{m=0}^{M-1} R\left(\frac{m}{M-1}\right),
\label{Eq8}
\end{equation}

where $M = 101$ denotes the number of uniformly spaced evaluation points. This metric encourages agents to complete more tasks in fewer steps, offering a holistic measure of task performance.

\subsection{Level 4: GUI Task Collaboration}
\label{sec:level_4}

Real-world task automation frequently requires agents to coordinate actions across multiple applications or environments, orchestrating complex workflows that involve heterogeneous interfaces and interdependent subtasks. To address such scenarios, task completion should extend beyond localized planning, requiring agents to develop a global perspective—tracking dependencies among applications, sequencing operations coherently, and managing cross-app information flow.
In this situation, agents must not only exhibit sophisticated long-horizon reasoning and planning abilities, but also handle practical challenges such as recovering from execution errors, coping with unexpected interface changes, and adapting to runtime variability in application responses. These factors collectively pose significant hurdles for contemporary GUI agents, making it a rigorous and realistic testbed for general-purpose automation intelligence.

Despite the critical role of collaboration and global reasoning in real-world workflows, existing benchmarks rarely address these aspects in a comprehensive and principled fashion.
Accordingly, we introduce \textit{L4-GUI Task Collaboration} as the fourth level of our benchmark, designed to systematically assess an agent's ability to use reasoning, collaboration, and adaptive automation across applications.

\renewcommand{\arraystretch}{1.2}
\begin{table*}[!t]
  \centering
  \caption{\textbf{Statistics of the evaluation data in MMBench-GUI.} Owing to the inherent restrictions of the iOS ecosystem, we were unable to include online tasks for iOS in L3\&L4. All other platforms are covered in full.}
  \small
  \begin{tabularx}{\textwidth}{>{\hsize=0.6\hsize}X*{6}{>{\centering\arraybackslash}X}>{\centering\arraybackslash}X}
    \toprule
     & Windows & MacOS & Linux & iOS & Android & Web & Overall \\
    \midrule
  
\multirow{6}{*}{\textbf{L1}} 
  & \multicolumn{7}{>{\columncolor{gray!20}}c}{\textcolor{gray}{L1 - Easy}} \\
 & 271 & 84  & 196  & 115  &  307 & 221  &  1194    \\
  & \multicolumn{7}{>{\columncolor{gray!20}}c}{\textcolor{gray}{L1 - Medium}} \\
  & 271 & 84  & 196  & 115  &  307 & 221  &  1194    \\     
  & \multicolumn{7}{>{\columncolor{gray!20}}c}{\textcolor{gray}{L1 - Hard}} \\
 & 271 & 84  & 196  & 115  &  307 & 221  &  1194    \\     

    \midrule
\multirow{4}{*}{\textbf{L2}} 
  & \multicolumn{7}{>{\columncolor{gray!20}}c}{\textcolor{gray}{L2 - Basic}} \\
  & 271 &  345  & 191  & 314  & 356  & 310  & 1787  \\
  & \multicolumn{7}{>{\columncolor{gray!20}}c}{\textcolor{gray}{L2 - Advanced}} \\
  &  272  & 346  & 196  & 330  & 335  & 308  & 1787  \\
\midrule
  \multirow{1}{*}{\textbf{L3}} 
 & 145 & 35  & 268  &  - & 116  & 155  &  719   \\

  \midrule
  \multirow{1}{*}{\textbf{L4}} 
  &  35 & 35  & 101  & -  & 30  &  47 &  248  \\

    \midrule
    \rowcolor{LakeBlue!15}
    \textbf{Total} & \textbf{1536} & \textbf{1013} & \textbf{1344} & \textbf{989} & \textbf{1758} & \textbf{1483} & \textbf{8123} \\

    \bottomrule
  \end{tabularx}
  
  \label{fig:statics}
\end{table*}

\textbf{Task Definition.} 
Extending the formulation above, L4 evaluates the agent’s ability to coordinate complex workflows involving multiple applications. 
The agent must generate and execute a sequence of actions that may interact with any application in the set $\mathcal{S}_m$, where $\mathcal{S}_m$ represents a subset of $k$ applications selected from the available pool, i.e., $\mathcal{S}_m \subseteq \{\texttt{App}_1, \texttt{App}_2, \ldots, \texttt{App}_N\}$ with $|\mathcal{S}_m| = k$, to accomplish a collaborative high-level task.
Formally, the agent-environment loop in Equation~\ref{Eq4} changes as follows:
\begin{equation}
\begin{split}
A_t, P_t &= \text{Agent}(\texttt{ins}, \mathbf{V}_t, \mathcal{H}_t, \mathcal{S}_m) \\
\mathbf{V}_{t+1} &= \text{Env}(A_t, P_t) \\
\mathcal{H}_{t+1} &= \{\mathcal{H}_t, (V_t, A_t, P_t)\}
\end{split}
\label{Eq9}
\end{equation}

Meanwhile, $\mathcal{H}_t$ now aggregates the interaction history across all relevant app environments. The process terminates when either the step limit is reached or a terminal action is predicted.

\textbf{Task Collection and Design.}
L4 tasks are designed as an extension of the single-app automation tasks in L3, with a primary focus on multi-application collaboration and information transfer across heterogeneous interfaces. For tasks in existing benchmarks that inherently involve multiple applications, we included them in our evaluation after a careful review of their availability and robustness. In addition, for those benchmarks lacking native multi-app workflows, we manually designed new tasks that explicitly require inter-app coordination. We also supplemented original multi-app tasks to further enrich the variety and complexity of cross-application scenarios.

A key design principle in constructing L4 tasks is to ensure that actions in one application provide necessary context or information for subsequent operations in another application. For example, in a representative macOS task, the agent is required to search online for the time and location of CVPR 2023 and then create a corresponding event in the Calendar app on the same date and month, but in the year 2090. To avoid issues related to time-sensitive information or changing event details, we decoupled the evaluation criteria from the actual event date, ensuring that the correctness of task completion is independent of the assessment time.

This systematic approach to task collection and design enables comprehensive evaluation of an agent’s ability to reason globally, manage inter-app dependencies, and execute complex workflows that mirror real-world user demands in multi-application environments. In the lower part of Figure~\ref{fig:level34_example}, we provide examples to illustrate how collaborative tasks involving two applications can be constructed.

\textbf{Evaluation Metrics.} We adopt the same evaluation metrics as in L3, i.e., SR and EQA. For both levels, the completion result is determined by verifying the final state and counting the number of steps taken, without the need to consider the individual states of multiple applications in $\mathcal{S}_m$.

\subsection{Benchmark Statistics}
\label{sec:statistics}

Table~\ref{fig:statics} enumerates the complete task inventory, 8123 distinct instances, broken down by operating platform, level, and difficulty band. Our benchmark has the following characteristics:
\begin{itemize}[leftmargin=2em]
    \item L1-GUI Content Understanding (3 × QA splits). Each of the six platforms contributes an identical triplet of 271/84/196/115/307/221 items (Windows $\rightarrow$ Web), yielding 1194 examples per difficulty (Easy, Medium, Hard) and 3582 in total. This symmetry ensures that any performance gap across the three difficulty tiers cannot be attributed to data imbalance.
    \item L2-GUI Element Grounding (Basic vs. Advanced). The grounding set is roughly 50\% larger than Level 1, with 1787 examples per split (Basic$=$Advanced). Note the deliberate platform skew: mobile platforms (iOS + Android = 686 or 38\%) receive more queries than desktop platforms, reflecting the higher UI diversity and screen density of mobile apps.
    \item L3-GUI Task Automation (single application). A compact but varied set of 719 trajectories focuses on long-horizon planning within one application.  Linux dominates (268 tasks) to capture the complexity of desktop productivity apps, while mobile splits are omitted for this level to avoid conflating OS diversity with task length.
    \item L4-GUI Task Collaboration (multiple applications). The hardest tier comprises 248 cross-application workflows. Although smaller, it intentionally spans all three desktop platforms and major mobile browsers (47 Web tasks, 30 Android tasks) to stress test memory hand-off and state persistence.
    \item Aggregate balance. Across the whole benchmark Windows (1536) and Android (1758) provide the two largest pools, but no single platform exceeds 22\% of the corpus, guarding against model over-specialisation. The progressive shrinkage, from 3582 (L1) to 248 (L4), mirrors the increasing cost and difficulty of annotation, while still offering enough samples (about 250) for a statistically meaningful evaluation in the top tier.
\end{itemize}

Overall, the benchmark delivers (1) platform diversity, (2) controlled difficulty gradation, and (3) a realistic taper in task count that matches real-world annotation effort, thereby enabling fine-grained diagnosis of GUI agent capabilities at every competence level.

\section{Benchmarking GUI Agent Baselines}
In this section, we evaluate a representative spectrum of contemporary VLM and LLM models, including both open-source and closed models, on the MMBench-GUI benchmark to provide a comprehensive portrait of current GUI-agent performance. MMBench-GUI supplies each method solely with screenshots and task descriptions, deliberately omitting auxiliary artifacts such as accessibility (A11y) trees and Set-of-Marks (SoM) data, thereby more closely mirroring real-world deployment scenarios. Since different models possess varying capabilities, the set of models evaluated is not entirely consistent across different levels of tasks. The details are as follows:

\textbf{L1\&L2}: Proprietary models: GPT-4o~\citep{hurst2024gpt}, Claude-3.7~\citep{anthropic2025claude37}, Qwen-Max-VL~\citep{Qwen-VL}. Open-sourced models: Qwen2.5 series~\citep{bai2025qwen25vl}, UI-TARS series~\citep{qin2025ui}, InternVL series~\citep{zhu2025internvl3}, Aguvis~\citep{xu2024aguvis}, ShowUI~\citep{lin2024showui}, UGround~\citep{gou2024navigating}, OS-Atlas~\citep{wu2024atlas}.

\textbf{L3\&L4}: Proprietary models: GPT-4o~\citep{hurst2024gpt}, Claude-3.7~\citep{anthropic2025claude37}. Open-sourced models: UI-TARS series~\citep{qin2025ui}, Qwen2.5-VL-72B~\citep{bai2025qwen25vl}, Aguvis~\citep{xu2024aguvis}, GPT-4o+UGround-V1-7B~\citep{gou2024navigating}, GPT-4o+UI-TARS-1.5-7B~\citep{qin2025ui}.

\subsection{Benchmarking details}
To ensure fairness, we evaluated all candidate models through a unified interface compatible with the OpenAI API protocol. Specifically, each model was deployed as an API-style service, and outputs were obtained by sending POST requests to the service endpoint along with the conversation input. For each model, we crafted both system and user prompts strictly based on official documentation or released code. For proprietary models, we designed detailed and effective prompts to elicit high-quality responses as faithfully as possible. Apart from model-specific settings, all other parameters, such as temperature and top-p, were kept consistent across evaluations.

During evaluation, the input and output processing pipeline was tailored to the requirements of each task level. For L1-GUI Content Understanding and L2-GUI Element Grounding, the input to the model comprised the GUI screenshot paired with either the relevant instruction or the question-options set. Model outputs were assessed using \texttt{exact-match} evaluation protocol, analogous to standard practices in grounding and QA tasks. However, given the variability in instruction-following abilities across different models, for example, the QA tasks in L1, we observed that some model outputs could not be reliably parsed. To address this, we implemented a hybrid parsing mechanism based on multiple regular expressions to robustly extract valid answers. In our codebase, we expose a customizable \texttt{parse\_function} for each method, enabling tailored post-processing strategies to accommodate the unique output formats of various models.

For L3-GUI Task Automation and L4-GUI Task Collaboration, evaluation focused solely on whether the agent successfully achieved the desired end state, without the need to interpret intermediate natural language outputs. Therefore, parsing functions were not required for these levels; instead, we compared the final state directly against predefined success criteria to determine task completion.

\subsection{Benchmark results on L1-GUI Content Understanding}

\renewcommand{\arraystretch}{1.2}
\begin{table*}[!t]
  \centering
  \caption{\textbf{Performance on L1-GUI Content Understanding.} `Overall' represents the aggregated score across all platforms, calculated as a weighted sum of individual platform scores. Here, the score of each platform is computed following Equation~\ref{eq:eq_4}, where $\alpha$ adjusts credit based on the number of candidate options for a question.}
  \small 
  \begin{tabularx}{\textwidth}{l*{6}{>{\centering\arraybackslash}X}>{\centering\arraybackslash}X}
    \toprule
    Model & Windows & MacOS & Linux & iOS & Android & Web & Overall \\
    \midrule
     \rowcolor{gray!20} \multicolumn{8}{c}{\textcolor{gray}{Easy Level}} \\
    GPT-4o~\citeyearpar{hurst2024gpt} & 62.47 & 67.89 & 62.38 & 58.52 & 56.41 & 58.51 & 60.16  \\
    Claude-3.5~\citeyearpar{claude35} &  41.34  &   50.04 & 41.61 & 42.03 & 38.96 & 41.79 &  41.54 \\
    Claude-3.7~\citeyearpar{anthropic2025claude37} & 34.66   &  49.05  & 39.37 & 42.76 & 37.45 & 40.80 &  39.08 \\
    Qwen-Max-VL~\citeyearpar{Qwen-VL} & \underline{69.05}   & 72.51   & 69.91 & \underline{70.82} & \underline{63.09} & 69.46 & \underline{68.15} \\
    Qwen2.5-VL-72B~\citeyearpar{bai2025qwen25vl} &  65.86  & \underline{75.23}   & \underline{73.02} & 67.24 & 58.09 & \underline{72.08} & 66.98 \\
    UI-TARS-72B-DPO~\citeyearpar{qin2025ui} & 41.59   &  28.52  & 35.16 & 31.08 & 52.25 & 35.33 &  40.18 \\
    InternVL3-72B~\citeyearpar{zhu2025internvl3} &  \textbf{74.67}  & \textbf{78.72} & \textbf{79.16} & \textbf{83.57} & \textbf{80.10} & \textbf{81.18} & \textbf{79.15}  \\

     \rowcolor{gray!20} \multicolumn{8}{c}{\textcolor{gray}{Medium Level}} \\
    GPT-4o~\citeyearpar{hurst2024gpt} &  56.33  & 63.13   & 59.70 & 54.06 & 57.69 & 54.98 & 57.24  \\
    Claude-3.5~\citeyearpar{claude35} &  39.28  &  47.63  & 45.97 & 44.57 & 42.03 & 34.33 & 41.26  \\
    Claude-3.7~\citeyearpar{anthropic2025claude37} &  39.34  & 39.23   & 42.28 & 39.45 & 36.05 & 36.17  & 38.39  \\
    Qwen-Max-VL~\citeyearpar{Qwen-VL} &  63.40  & \underline{73.85}   & 66.90 & \underline{68.02} & 63.66 & 64.59 & 65.44 \\
    Qwen2.5-VL-72B~\citeyearpar{bai2025qwen25vl} &  \underline{66.29}  &  72.73  & \underline{72.63} & 59.27 & \underline{66.24} & \underline{68.24} & \underline{67.45} \\
    UI-TARS-72B-DPO~\citeyearpar{qin2025ui} & 38.83   &  41.60  & 37.14 & 41.72 & 54.74 & 31.55 & 41.77  \\
    InternVL3-72B~\citeyearpar{zhu2025internvl3} &  \textbf{71.46}  & \textbf{78.58}   &\textbf{79.88} & \textbf{78.43} & \textbf{81.36} &\textbf{78.67} & \textbf{77.89} \\

     \rowcolor{gray!20} \multicolumn{8}{c}{\textcolor{gray}{Hard Level}} \\
    GPT-4o~\citeyearpar{hurst2024gpt} &  60.69  &  60.38  & 52.42 & 45.27 & 50.93 & 50.83 &  53.49 \\
    Claude-3.5~\citeyearpar{claude35} & 37.40   &  42.70  & 34.07 & 40.86 & 36.96 & 38.11 & 37.55  \\
    Claude-3.7~\citeyearpar{anthropic2025claude37} &  32.99  &  34.48  & 31.97 & 39.20 & 36.99 & 38.92 &  35.65 \\
    Qwen-Max-VL~\citeyearpar{Qwen-VL} &  66.64  &  67.59  & 65.80 & \underline{60.23} & \underline{58.78} & 65.34 & 63.69 \\
    Qwen2.5-VL-72B~\citeyearpar{bai2025qwen25vl} &  \underline{70.68}  &  \underline{68.91}  & \underline{70.98} & 57.59 & 53.94 & \underline{68.10} & \underline{64.56} \\
    UI-TARS-72B-DPO~\citeyearpar{qin2025ui} &  31.48  &  35.87  & 24.19 & 36.33 & 58.13 & 19.94 & 35.78  \\
    InternVL3-72B~\citeyearpar{zhu2025internvl3} & \textbf{75.08}   & \textbf{77.44}   & \textbf{76.19} &\textbf{70.37}& \textbf{75.73} & \textbf{78.11} &  \textbf{75.70} \\
    \bottomrule
  \end{tabularx}
  \label{tab:understanding}
\end{table*}

Table~\ref{tab:understanding} summarizes the performance of all evaluated models on the GUI Understanding task (L1) across three difficulty levels (Easy, Medium, Hard) and six platforms (Windows, MacOS, Linux, iOS, Android, Web), as well as the overall average.
Across all settings, InternVL3-72B consistently achieves the highest scores, outperforming all other models on every platform and difficulty tier. Qwen2.5-VL-72B and Qwen-Max-VL generally rank just below InternVL3-72B. GPT-4o exhibits moderate performance, while the Claude variants (3.5 and 3.7) and UI-TARS-72B-DPO perform less favorably across all settings.

Several consistent trends emerge from the results:
\begin{itemize}[leftmargin=2em]
    \item Difficulty effect: Model performance decreases as task difficulty increases, with scores on the Easy level always exceeding those on Medium and Hard levels.
    \item Cross-platform variability: For most models, macOS and Linux yield slightly higher scores, whereas Android and Web present greater variability and, in some cases, lower accuracy, indicating additional platform-specific challenges.
    \item Model ranking and robustness: InternVL3-72B maintains its leading position across all difficulty tiers (overall: 79.2\%, 77.9\%, and 75.7\% on Easy, Medium, and Hard, respectively) and shows the smallest decline in performance as difficulty increases. Qwen2.5-VL-72B consistently ranks second, while GPT-4o experiences a sharper drop on harder items. The Claude variants and UI-TARS-72B-DPO show both lower accuracy and limited robustness across difficulty levels.
\end{itemize}

Overall, these results demonstrate clear differences in model capabilities on GUI content understanding tasks, providing a solid quantitative basis for the in-depth analysis presented in the next section.

\subsection{Benchmark results on L2-GUI Element Grounding}

\renewcommand{\arraystretch}{1.2} 
\begin{table*}[!t]
  \centering
  \caption{\textbf{Performance on the L2-GUI Element Grounding.} “Adv.” stands for advanced, while “Avg.” refers to the weighted average of all results in a row, where the weights correspond to the proportion of tasks for each platform and mode relative to the total number of tasks.}
  \small
  \begin{tabularx}{\textwidth}{l*{13}{>{\centering\arraybackslash}X}}
    \toprule
    \multirow{2}{*}{\textbf{Model}} 
      & \multicolumn{2}{c}{Windows} 
      & \multicolumn{2}{c}{MacOS} 
      & \multicolumn{2}{c}{Linux} 
      & \multicolumn{2}{c}{iOS} 
      & \multicolumn{2}{c}{Android} 
      & \multicolumn{2}{c}{Web} 
      & \multirow{2}{*}{Avg}\\
    \cmidrule(r){2-3}
    \cmidrule(r){4-5}
    \cmidrule(r){6-7}
    \cmidrule(r){8-9}
    \cmidrule(r){10-11}
    \cmidrule(r){12-13}
      & Basic & Adv. & Basic & Adv. & Basic & Adv. & Basic & Adv. & Basic & Adv. & Basic & Adv. & \\
    \midrule
    GPT-4o~\citeyearpar{hurst2024gpt} & 1.48 &  1.10  & 8.69 & 4.34 & 1.05 & 1.02 & 5.10 & 3.33 & 2.53 & 1.41 & 3.23 & 2.92 & 2.87\\
    Claude-3.7~\citeyearpar{anthropic2025claude37} & 1.48   & 0.74   & 12.46 & 7.51 & 1.05 & 0.00 & 13.69 & 10.61 & 1.40 & 1.40 & 3.23 &2.27 & 4.66 \\
    Qwen-Max-VL~\citeyearpar{Qwen-VL} &  43.91  &  36.76  & 58.84 & \underline{56.07} & 53.93 & 30.10 & 77.39 & 59.09 & 79.49 & 70.14 & 74.84 & 58.77 & 58.03 \\
    Aguvis-7B-720P~\citeyearpar{xu2024aguvis} &  37.27  &  21.69  & 48.12 & 33.27 & 33.51 & 25.00 & 67.52 & 65.15 & 60.96 & 50.99 & 61.61 & 45.45 & 45.66\\
    ShowUI-2B~\citeyearpar{xu2024aguvis} &  9.23  &  4.41  & 24.06 & 10.40 & 25.13 & 11.73 &28.98  & 19.70 & 17.42 & 8.73 & 22.90 & 12.66 & 15.96 \\
    OS-Atlas-Base-7B~\citeyearpar{wu2024atlas}  &  36.90  &  18.75  & 44.35 & 21.68 & 31.41 & 13.27  & 74.84 & 48.79  & 69.6 & 46.76  & 61.29 & 35.39 & 41.42\\
    UGround-V1-7B~\citeyearpar{gou2024navigating} &  66.79  &  38.97  & 71.30 & 48.55 & 56.54 & 31.12 & \underline{92.68} & 70.91 & \textbf{93.54} & 70.99 & \underline{88.71} & 64.61 & 65.68 \\
    InternVL3-72B~\citeyearpar{zhu2025internvl3} &  \underline{70.11}  & \underline{42.64}   & \underline{75.65}& 52.31 & 59.16 & \underline{41.33} & \textbf{93.63} & \underline{80.61} & 92.70 & \underline{78.59} & \textbf{90.65} & \underline{65.91} & \underline{72.20}\\
    Qwen2.5-VL-72B~\citeyearpar{bai2025qwen25vl} &  55.72  &  33.82  & 49.86 & 30.06 & 40.31 & 20.92 & 56.05 & 28.18 & 55.62 & 25.35 & 68.39 & 45.78 & 41.83\\
    Qwen2.5-VL-7B~\citeyearpar{bai2025qwen25vl} &  31.37  &  16.54  & 31.30 & 21.97 & 21.47 & 12.24 & 66.56 & 55.15 & 35.11 & 35.21 & 40.32 & 32.47 & 33.85 \\
    UI-TARS-1.5-7B~\citeyearpar{qin2025ui} &  68.27 &  38.97  & 68.99 & 44.51 & \underline{64.40} & 37.76 & 88.54 & 69.39 & 90.45 & 69.29 & 80.97 & 56.49  & 64.32 \\
    UI-TARS-72B-DPO~\citeyearpar{qin2025ui} &  \textbf{78.60}  & \textbf{51.84}  & \textbf{80.29} & \textbf{62.72} & \textbf{68.59} &\textbf{51.53} & 90.76 &\textbf{81.21}& \underline{92.98} & \textbf{80.00} & 88.06& \textbf{68.51} & \textbf{74.25} \\
    \bottomrule
  \end{tabularx}
  \vspace{-3mm}
  \label{tab:grounding}
\end{table*}

Table~\ref{tab:grounding} reports the results of all evaluated models on the L2 task, including both Basic and Advanced instructions, across six platforms. We can summarize the following:

\begin{itemize}[leftmargin=2em]
    \item Significant variation is observed among models. GPT-4o and Claude-3.7 exhibit extremely limited grounding ability, with scores consistently near zero across all platforms and instruction types. In contrast, open-source models such as UI-TARS-72B-DPO, InternVL3-72B, UGround-V1-7B, and Qwen2.5-VL-72B achieve substantially higher scores.
    \item The best-performing models (UI-TARS-72B-DPO and InternVL3-72B) demonstrate both high overall averages (74.25\% and 72.20\%, respectively) and strong cross-platform consistency. For example, UI-TARS-72B-DPO achieves average scores above 80\% for MacOS, Android, and Web in the Basic setting, while maintaining robust performance on iOS (62.72\%) and Linux (68.59\%). InternVL3-72B similarly shows strong results across all platforms.
    \item A clear platform-dependent pattern emerges. For most high-performing models, grounding accuracy is generally higher on mobile (iOS, Android) and web platforms, with somewhat lower scores on desktop environments (Windows, MacOS, Linux). For instance, UI-TARS-72B-DPO achieves 93.54\% on Android (Basic), 88.71\% on Web (Basic), but comparatively lower scores on Windows (78.60\%, Basic) and Linux (68.59\%, Basic).
    \item Model performance generally drops from Basic to Advanced instruction types. While top models maintain a high level on both, their scores under Advanced instructions are consistently lower than under Basic instructions, suggesting increased difficulty with more abstract or functional cues.
\end{itemize}

In summary, these results indicate wide gaps in GUI element grounding capabilities among current models, as well as persistent platform and instruction-type differences. We will make deeper analysis in the following sections.

\subsection{Benchmark results on L3-GUI Task Automation and L4-GUI Task Collaboration}

\renewcommand{\arraystretch}{1.3} 
\begin{table*}[!t]
  \centering
  \caption{\textbf{Evaluation result of L3-GUI Task Automation.} Values in \textbf{bold} indicate the highest score within each group; \underline{underlined} values indicate the second highest.}
  \small
  \begin{tabularx}{\textwidth}{l*{12}{>{\centering\arraybackslash}X}}
    \toprule
    \multirow{2}{*}{\textbf{Model}} 
      & \multicolumn{2}{c}{Windows} 
      & \multicolumn{2}{c}{MacOS} 
      & \multicolumn{2}{c}{Linux} 
      & \multicolumn{2}{c}{Android} 
      & \multicolumn{2}{c}{Web} 
      & \multicolumn{2}{c}{Avg}\\
    \cmidrule(r){2-3}
    \cmidrule(r){4-5}
    \cmidrule(r){6-7}
    \cmidrule(r){8-9}
    \cmidrule(r){10-11}
    \cmidrule(r){12-13}
      & SR & EQA & SR & EQA & SR & EQA & SR & EQA & SR & EQA & SR & EQA \\
    \midrule
    \rowcolor{gray!20} \multicolumn{13}{c}{\textcolor{gray}{Max Step=15}} \\
    GPT-4o~\citeyearpar{hurst2024gpt} &  5.56  &  3.27  & 0.00 & 0.00 & 6.83 & 4.35 & 18.97 & 8.93 & 1.94 & 1.53 & 7.14 & 4.05 \\
    Claude-3.7~\citeyearpar{anthropic2025claude37} & 7.09 & 4.28 & \underline{8.57} & 2.76 & 7.43 & 4.20 & 11.21 & 3.49 & 1.94 & 1.46 & 6.84 & 3.44 \\
    Aguvis-72B~\citeyearpar{xu2024aguvis} & 4.14 & 2.02  & 0.00 & 0.00 & 3.09 & 1.63 & 18.10 & 10.78 & 9.03 & 3.75 & 6.85 & 3.56 \\
    UI-TARS-1.5-7B~\citeyearpar{qin2025ui} & 11.08 & 5.98 & \textbf{11.43} & \underline{6.58} & \underline{26.51} & \underline{18.65} & 30.17 & 17.93 & 12.26 & 6.98 & 20.18 & \underline{12.87} \\
    UI-TARS-72B-DPO~\citeyearpar{qin2025ui} &  11.08  &  5.44 & \textbf{11.43} & \textbf{7.79} & \textbf{30.31} & \textbf{18.91} & \textbf{43.10} & \textbf{26.62} & 10.32 & 6.94 & \textbf{23.27} & \textbf{14.31} \\
    Qwen2.5-VL-72B~\citeyearpar{bai2025qwen25vl} &   11.77  & 7.18    &  2.86   &  2.01    &   9.80   &  5.37   &   16.37   &   9.77    &   15.58   &   9.92    &  12.17    &  7.26      \\
    GPT-4o + UGround-V1-7B~\citeyearpar{gou2024navigating}  & \underline{13.10}   &  \textbf{8.11}  & 2.86 & 1.00 & 16.13 & 8.69 & \underline{34.48} & \underline{21.14} & \textbf{23.23} & \textbf{16.69} & 19.36 & 11.93 \\
    GPT-4o + UI-TARS-1.5-7B~\citeyearpar{qin2025ui} &  \textbf{14.52}  &  \underline{6.76}  & 2.86 & 0.91 &20.23 &11.12& 33.62 & 15.17 & \underline{22.58} & \underline{14.65} & \underline{20.90} & 11.16 \\
    \rowcolor{gray!20} \multicolumn{13}{c}{\textcolor{gray}{Max Step=50}} \\
    GPT-4o~\citeyearpar{hurst2024gpt} &  3.49  &  2.26  & 2.86 & 1.65 & 11.64 & 9.05 & 21.55 & 10.81 & 3.23 & 2.24 & 9.35 & 6.13 \\
    Claude-3.7~\citeyearpar{anthropic2025claude37} & 6.40 & 4.03 & \textbf{11.43} & 4.23 & 10.28 & 6.25 & 11.21 & 3.62 & 2.58 & 2.11 & 8.04 & 4.39 \\
    Aguvis-72B~\citeyearpar{xu2024aguvis} & 3.49 & 1.63 & 0.00 & 0.00 &4.21  & 2.04 & 19.83 & 14.67 & 8.39 & 3.10 & 7.28 & 4.12 \\
    UI-TARS-1.5-7B~\citeyearpar{qin2025ui} & 15.86 &  11.29  & \textbf{11.43}  & \underline{7.03} & \underline{29.82} & \underline{21.26} & 31.58 & 22.15 & 14.19 & 9.22 & 23.02 & 16.10 \\
    UI-TARS-72B-DPO~\citeyearpar{qin2025ui} &  17.93  & 11.84   &  \textbf{11.43} & \textbf{8.38}  & \textbf{31.38} & \textbf{25.44} & \underline{45.69} & \underline{35.22} & 9.68 & 7.53 & \underline{25.33} & \textbf{19.58} \\
    Qwen2.5-VL-72B~\citeyearpar{bai2025qwen25vl} &  9.66   &  6.86   &  5.71   &  3.96    &   10.63   &  7.85   &   27.59   &   21.80    &   14.38   &   9.74    &   13.74   &   10.12     \\
    GPT-4o + UGround-V1-7B~\citeyearpar{gou2024navigating}  & \underline{20.73}   & \underline{11.89}   & 5.71 & 3.18 & 19.48 & 10.91 & \textbf{47.41} & \textbf{37.19} & \textbf{26.45} & \textbf{22.50} & 25.07 & 17.50 \\
    GPT-4o + UI-TARS-1.5-7B~\citeyearpar{gou2024navigating} & \textbf{26.21} & \textbf{17.28} &  \underline{8.57}  & 5.01  &22.85  & 13.82 & 42.24 & 33.10 & \underline{25.81} & \underline{20.72} & \textbf{26.60} & \underline{18.69} \\
    \bottomrule
  \end{tabularx}

  \label{tab:singleapp-automation}
\end{table*}

\renewcommand{\arraystretch}{1.3} 
\begin{table*}[!t]
  \centering
  \caption{\textbf{Evaluation result of L4-GUI Task Collaboration.} "-" represents that these model's action space can't handle browser tab switch situation, so we don't test them.}
  \small
  \begin{tabularx}{\textwidth}{l*{13}{>{\centering\arraybackslash}X}}
    \toprule
    \multirow{2}{*}{\textbf{Model}} 
      & \multicolumn{2}{c}{Windows} 
      & \multicolumn{2}{c}{MacOS} 
      & \multicolumn{2}{c}{Linux} 
      & \multicolumn{2}{c}{Android} 
      & \multicolumn{2}{c}{Web} 
      & \multicolumn{2}{c}{Avg}\\
    \cmidrule(r){2-3}
    \cmidrule(r){4-5}
    \cmidrule(r){6-7}
    \cmidrule(r){8-9}
    \cmidrule(r){10-11}
    \cmidrule(r){12-13}
      & SR & EQA & SR & EQA & SR & EQA & SR & EQA & SR & EQA  & SR & EQA \\
    \midrule
    \rowcolor{gray!20} \multicolumn{13}{c}{\textcolor{gray}{Max Step=15}} \\
    GPT-4o~\citeyearpar{hurst2024gpt} & 7.49 & \underline{5.90} & 0.00  & 0.00 & 3.50 & 2.41& 0.00 & 0.00 & \underline{2.13} & 0.10 & 2.85 & 1.80  \\
    Claude-3.7~\citeyearpar{anthropic2025claude37} & 3.57  & 1.61 & \underline{2.86} & \textbf{2.01} & \underline{7.32} & \underline{4.76} & 0.00 & 0.00 & \underline{2.13} & 0.03 & 4.30 & 2.46 \\
    Aguvis-72B~\citeyearpar{xu2024aguvis} & 3.21 & 3.01 & 0.00 & 0.00 & 1.62 & 0.37 & 3.33 & 3.15 & - & - & 1.50 & 0.94 \\
    UI-TARS-1.5-7B~\citeyearpar{qin2025ui} &  3.21  &  3.01  & \underline{2.86} & 0.82 & 4.95 & 3.98 & 6.67 & 6.58 & - & - & 3.68 & 2.96 \\
    UI-TARS-72B-DPO~\citeyearpar{qin2025ui} & 3.21 & 3.05   & \textbf{5.71} & \underline{1.47} & \textbf{7.46}  & \textbf{5.87} & \underline{10.00} & \underline{9.64} & - & - & \underline{5.53} & \underline{4.20}  \\
    Qwen2.5-VL-72B~\citeyearpar{bai2025qwen25vl} &  6.24   &  4.20   &  0.00   &   0.00   &   2.53   &  1.65   &   6.67   &  6.08     &   -   &   -    &  3.35    & 2.47       \\
    GPT-4o + UGround-V1-7B~\citeyearpar{gou2024navigating}  & \underline{9.27} & 5.12 & 0.00 & 0.00 & 3.60 & 2.51 & 3.33 & 3.22 & \textbf{4.26} & \underline{0.61} & 3.94 & 2.23 \\
    GPT-4o + UI-TARS-1.5-7B~\citeyearpar{qin2025ui} & \textbf{12.30}  & \textbf{6.36}   & 0.00 & 0.00 &  5.58& 3.82 & \textbf{23.33} & \textbf{21.12} & \textbf{4.26} & \textbf{0.68} & \textbf{7.6} &  \textbf{5.13} \\
    \rowcolor{gray!20} \multicolumn{13}{c}{\textcolor{gray}{Max Step=50}} \\
    GPT-4o~\citeyearpar{hurst2024gpt} &  6.24  &  4.98  & 0.00 & 0.00 & 5.94 & 5.35 & 0.00 & 0.00 & \underline{2.13} & \underline{1.52} & 3.68 & 3.16 \\
    Claude-3.7~\citeyearpar{anthropic2025claude37} & 6.24   & 4.34   & \underline{2.86} & \underline{2.03} & \textbf{9.30} & \textbf{7.35}  & 0.00 & 0.00 & \underline{2.13} & 0.04 & 5.47 & 3.90 \\
    Aguvis-72B~\citeyearpar{xu2024aguvis} &  3.21  &  3.14  & 0.00  & 0.00  & 1.62 & 0.37 & 6.67 & 6.42 & - & - & 1.91 & 1.36 \\
    UI-TARS-1.5-7B~\citeyearpar{qin2025ui} &  6.24  &  6.00  & \underline{2.86} & 0.89 & 7.63 & 5.50 & 13.33 & \underline{13.07} & - & - & 6.00 & 4.78 \\
    UI-TARS-72B-DPO~\citeyearpar{qin2025ui} & \underline{9.27}   &  \underline{6.22}  & \textbf{5.71} & \textbf{2.27} & \underline{8.45} & \underline{7.19} & \underline{20.00} & 11.78 & - & - & \underline{7.96} & \underline{5.55} \\
    Qwen2.5-VL-72B~\citeyearpar{bai2025qwen25vl} &  6.24   &  5.23  &  0.00   &   0.00   &  1.62    & 1.31    &    6.67  &  6.51     &  -    &   -    &  2.90    & 2.82       \\
    GPT-4o + UGround-V1-7B~\citeyearpar{gou2024navigating}  & \underline{9.27} &  5.03  & 0.00 & 0.00 & 5.48 & 3.75 & 6.67 & 6.40 & 0.00 & 0.00 & 4.31 & 2.99 \\
    GPT-4o + UI-TARS-1.5-7B~\citeyearpar{gou2024navigating} & \textbf{12.30} &  \textbf{6.84}  & 2.86 & 0.95 & 7.46 & 5.59 & \textbf{23.33} & \textbf{21.65} & \textbf{4.26} & \textbf{2.02} & \textbf{8.78} & \textbf{6.37}  \\
    \bottomrule
  \end{tabularx}

  \label{tab:multiapp-automation}
  \vspace{-3mm}
\end{table*}

Tables~\ref{tab:singleapp-automation} and~\ref{tab:multiapp-automation} report the results for all models on single-app (L3) and multi-app (L4) GUI automation tasks under 15 and 50 steps. For all models, we employed a unified evaluation pipeline, using standardized prompts and action spaces for general-purpose models, and official configurations for GUI-specific agents.

For L3 tasks, overall performance is limited across all models and platforms. The best-performing method, GPT-4o + UI-TARS-1.5-7B, achieves an average SR of 26.60\% , while most other models remain below 20\%. The EQA scores follow a similar trend. Among GUI-specific models, UI-TARS-72B-DPO shows the best overall SR and EQA, particularly outperforming other agents on Linux and Android.
Notably, language-centric models, that is, GPT-4o and Claude-3.7, perform less favorably across platforms and metrics. However, combining general-purpose models with GUI-specific grounders, such as UGround or UI-TARS, consistently boosts performance; for instance, GPT-4o alone achieves 6.13\% SR, but this rises to over 17\% with planner+grounder variants.

For L4, model success rates are considerably lower. The top method (GPT-4o + UI-TARS-1.5-7B) achieves only 8.78\% average SR, and most models fall below 6\%. This substantial drop compared to L3 underscores the increased difficulty in execution of cross-application tasks.

Increasing the maximum allowed steps from 15 to 50 improves SR and EQA values for all models and settings, but the overall task completion rates remain low, indicating that simply allowing longer action sequences does not fully address the challenges. This suggests that even with greater execution flexibility, many agents still struggle with effective long-horizon planning and multi-step task execution.

Platform-wise, Android and Web tend to yield higher SR and EQA for top-performing models (e.g., GPT-4o + UI-TARS-1.5-7B achieves SR/EQA of 33.10\%/25.81\% on Android and 20.72\%/20.72\% on Web, compared to 26.21\%/17.28\% on Windows and 8.57\%/5.01\% on macOS), while desktop environments, especially macOS, generally show lower results.

The results from Tables~\ref{tab:singleapp-automation} and~\ref{tab:multiapp-automation} highlight both the effectiveness of combining planning and grounding for L3 tasks and the substantial gap in agent performance when moving to L4 scenarios, especially under long-horizon and multi-step conditions.

\section{Analysis and discussion}
In this section, we conduct an in-depth analysis to delve into the underlying causes and implications reflected in our benchmark results. Our investigation is structured around three primary dimensions: platform, task, and model, and adheres to a single-variable control principle to ensure the validity of our comparisons. Through systematic examination and post-processing of the empirical results along these axes, we distill a series of actionable findings that reveal the fundamental bottlenecks currently constraining agent performance. These findings not only elucidate the essential challenges facing contemporary GUI agents, but also offer valuable guidance for future research and development in this domain.

\textbf{Finding 1: General-purpose language models excel at task decomposition, planning, and self-reflection but struggle with fine-grained visual interactions.}

\begin{figure*}[!t]
    \centering
    \includegraphics[width=\linewidth]{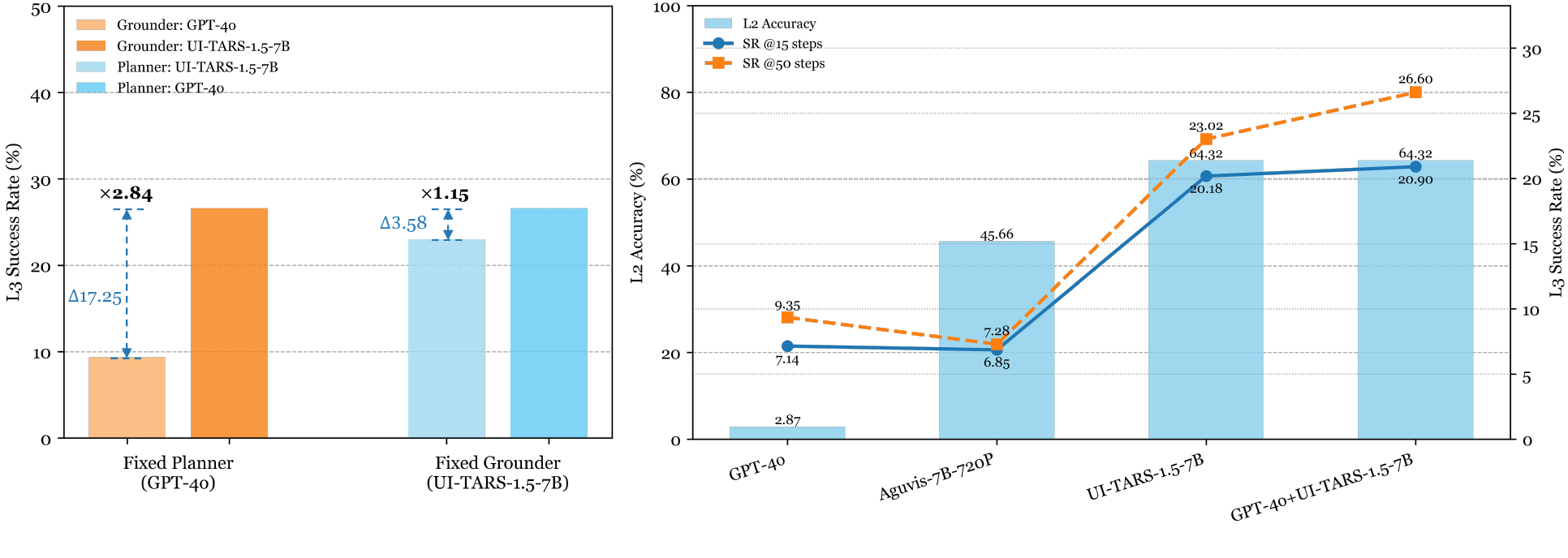}
    \caption{\textbf{Left:} Demonstrates the relative contribution of visual grounding versus planning in driving performance gains under current conditions. We consider two experimental conditions—fixing the planner while varying the grounder, and vice versa—and examine how different combinations affect task success rate. Similar color hues denote groups with the same fixed planner or grounder. \textbf{Right:} Task success grows roughly linearly with visual-grounding accuracy. General-purpose language models are virtually “blind” at the L2 grounding stage, which drives their L3 automation success rate (SR) sharply down. Plugging in a dedicated visual grounder restores precise perception and, in turn, lifts SR dramatically—highlighting fine-grained grounding as the principal bottleneck.}
    \vspace{-2mm}
    \label{fig:Importance_of_grounding}
\end{figure*}

Across different model categories, general-purpose language models, exemplified by GPT-4o and Claude, demonstrate pronounced limitations in fine-grained GUI tasks. As shown in Table~\ref{tab:grounding} and the right part of Figure~\ref{fig:Importance_of_grounding}, their average scores in L2 are merely 2.87 for GPT-4o and 4.66 for Claude-3.7, in contrast to the specialized visual grounding model UGround-V1-7B, which achieves a score of 65.68\%. This discrepancy underscores a key limitation: general-purpose models inherently lack the capacity for accurate perception and localization of UI components. A similar trend emerges in L3 tasks. For instance, GPT-4o alone achieves success rates (SR) of only 4.05\%/6.13\% in single-app automation scenarios (Max Step = 15/50, see Table~\ref{tab:singleapp-automation}). However, when paired with domain-specific grounding modules such as UGround-V1-7B or UI-TARS-1.5-7B, the SR of GPT-4o rises substantially to 11.93\%/17.50\%. This  phenomenon suggests that specialized perception modules can effectively compensate for the perceptual shortcomings of general-purpose LLMs.

Beyond the two direct strategies, namely, incorporating auxiliary localization modules during training and increasing the amount of fine-grained perceptual data, a more fundamental and forward-looking direction lies in embracing a modular architecture. This approach enables the model to dynamically interface with external modules based on its own capability gaps (e.g., visual grounding), effectively allowing for targeted augmentation through specialized ``external agents”. This architecture not only compensates for inherent deficiencies but also promotes a flexible, cooperative paradigm in which general-purpose models can be extended and adapted to complex GUI automation tasks.

\textbf{Finding 2: Accurate visual grounding significantly determines the success rate of GUI task execution.}

The full decision-making pipeline of a GUI agent can be abstracted into three stages: perceive accurately $\Rightarrow$ reason properly $\Rightarrow$ act precisely. If the first step (element localization) fails, subsequent planning and reasoning, no matter how advanced, are unlikely to compensate. To examine the critical role of localization, we systematically assessed its impact on downstream automation tasks (L3 and L4), and conversely, investigated whether enhanced planning alone could offset poor visual grounding.

We designed two complementary experimental setups as shown in the left part of Figure~\ref{fig:Importance_of_grounding}: (1) fixing the planner while incrementally improving the grounder, and (2) fixing the grounder while varying the planner. Correlation analyses revealed a clear pattern: with the same planner, improving localization alone led to a ~2.8× ($\Delta=17.25$) increase in SR. In contrast, when localization performance remained roughly constant, replacing the planner with a stronger VLM yielded marginal returns (1.15×, $\Delta=3.58$). These results lead to a clear conclusion: visual grounding is the primary performance bottleneck. Gains from improved localization are nearly linear, whereas once the agent "sees well enough," the marginal utility of enhancing its reasoning diminishes.

This finding underscores that, at the current stage, the most leverage-efficient breakthrough for improving GUI task automation lies in advancing high-precision, cross-platform visual localization capabilities. As also suggested by Finding 1, within a modular architecture, the visual grounder should be treated as the first and most critical plug-in component. Ensuring its reliability provides a solid foundation upon which LLM-based planning, long-range memory, and reflection mechanisms can be effectively layered.

\textbf{Finding 3: Efficiency, including step minimization and early stopping, is a critical yet underexplored dimension of GUI agent performance.
}

The introduction of the EQA metric enables us to move beyond evaluating whether an agent simply completes a task, by shifting attention to how efficiently the task is accomplished. This novel perspective facilitates deeper insights through a more fine-grained analysis of agent behavior.

We additionally compute two derived metrics, $\frac{EQA}{SR}$ and $SR-EQA$, to facilitate a more comprehensive analysis. Based on the definition of the EQA and SR, we further reformulate them as:
\begin{equation}
    EQA=\frac{1}{N}\sum_{i \in \mathcal{C}} (1-u_i), \ \ SR=\frac{|\mathcal{C}|}{N},
\end{equation}
where $\mathcal{C}$ denotes the set of all successfully completed tasks, and $u_i = \frac{T_i}{T_{max}} \in (0,1]$ represents the normalized completion step of task $i$ within the global step budget. From this, $\frac{EQA}{SR}$ and $SR-EQA$ can be derived as:
\begin{equation}
    \frac{EQA}{SR} = \frac{1}{\mathcal{|C|}}\sum_{i \in \mathcal{C}} (1-u_i) = 1 - \frac{1}{\mathcal{|C|}}\sum_{i \in \mathcal{C}} u_i,
\end{equation}
\begin{equation}
    SR-EQA = \frac{1}{N}\sum_{i \in \mathcal{C}} u_i = SR \times (\frac{1}{\mathcal{|C|}}\sum_{i \in \mathcal{C}} u_i),
\end{equation}
where $\frac{1}{\mathcal{|C|}}\sum_{i \in \mathcal{C}} u_i$ denotes the average steps in which a task is completed. 

Therefore, $\frac{EQA}{SR}$ has an intuitive physical interpretation: it reflects the average remaining steps per successful task. Its upper bound is 1, which corresponds to the idealized case where all successful tasks are completed almost immediately (i.e., at the first step). Conversely, its lower bound is 0, indicating that all successful completions occur only at the very end of the allowed budget. $\frac{EQA}{SR}$ quantifies how many steps, on average, are consumed before successful completion. Meanwhile, $SR-EQA$ also has an intuitive physical interpretation: it is approximately proportional to the total normalized time consumed across all successful tasks, and can be interpreted as a “redundant step bill”. A larger difference between EQA and SR implies a greater average normalized completion time $u_i$ for the successful set, meaning that tasks tend to be completed closer to the end of the budget—i.e., with more redundant steps.
Conversely, a smaller difference (approaching zero) indicates that most successful tasks are completed early, near the beginning of the budget, suggesting minimal or no redundancy. Thus, the magnitude of the gap between EQA and SR effectively captures how “wasteful” the agent is, even among the tasks it completes.

\begin{figure*}[!t]
    \centering
    \includegraphics[width=\linewidth]{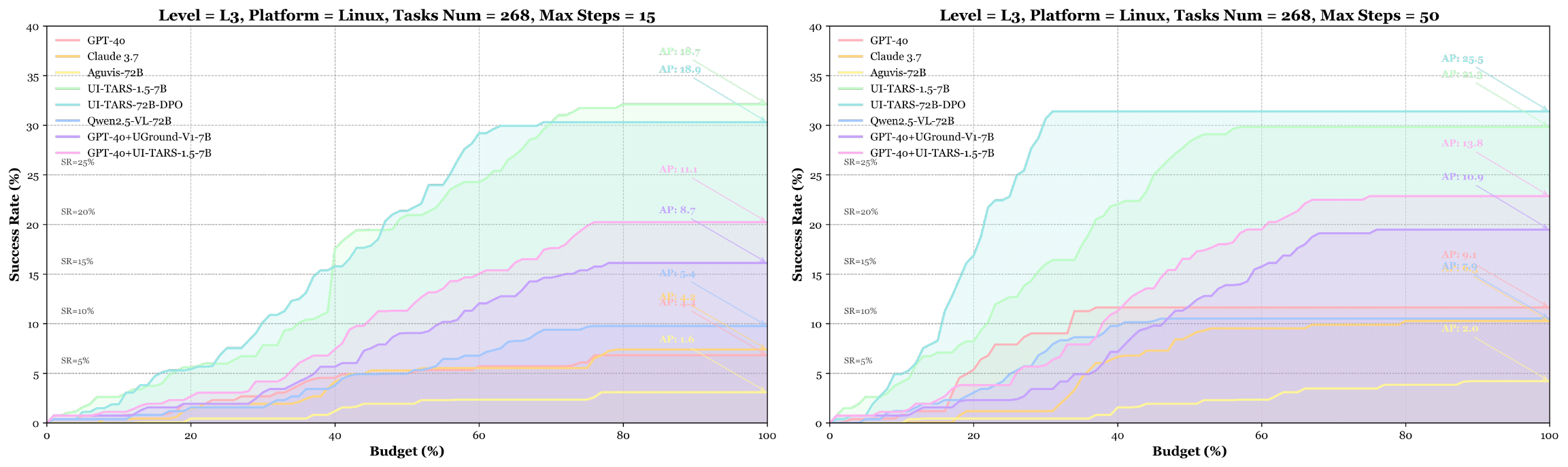}
    \caption{\textbf{EQA visualization across different models under L3 for different allowed steps.} As discussed in Section~\ref{sec:level_3}, EQA reflects a combination of task completion and efficiency (i.e., the number of steps used upon completion). In practice, we compute it by interpolating both the step budget and the success rate (SR) 100 times. The area under the curve formed by these interpolated SR values yields the final EQA score.}
    \vspace{-2mm}
    \label{fig:EQA_visualization}
\end{figure*}
\renewcommand{\arraystretch}{1.1} 
\begin{table*}[t]
  \centering
  \small
  \begin{tabularx}{\textwidth}{l*{9}{>{\centering\arraybackslash}X}}
    \toprule
    \multirow{1}{*}{\textbf{Model}} 
      & \multicolumn{1}{c}{$\Delta\text{SR}$}
      & \multicolumn{1}{c}{$\Delta\text{EQA}$} 
      & \multicolumn{1}{c}{$\text{EQ}_{15}^1$} 
      & \multicolumn{1}{c}{$\text{EQ}_{50}^1$} 
      & \multicolumn{1}{c}{$\text{EQ}_{15}^2$} 
      & \multicolumn{1}{c}{$\text{EQ}_{50}^2$} 
      & \multicolumn{1}{c}{$\Delta \text{EQ}^1$}
      & \multicolumn{1}{c}{$\Delta \text{EQ}^2$} \\

    \midrule
    GPT-4o & 	2.21&	2.08&	0.567 &	0.656 & 3.09 & 3.22 &	0.088  & 0.13 \\
    Claude-3.7 & 	1.2	&0.95&	0.503 &	0.546 & 3.40 & 3.65  &	0.043 & 0.25 \\
    Aguvis-72B & 	0.43	&0.56	&0.520& 	0.566 & 3.29 & 3.16 &	0.046 & -0.13\\
    UI-TARS-1.5-7B & 2.84&	3.23&	0.638 &	0.699 & 7.31 & 6.92 &	0.062 & -0.39 \\
    UI-TARS-72B-DPO &	2.06	&5.27	&0.615& 	0.773 & 8.96 & 5.75 & 	0.158   & -3.21 \\
    Qwen2.5-VL-72B & 	1.57&	2.86	&0.597 &	0.737 & 4.91& 3.62&	0.140 & -1.29\\
    GPT-4o+UGround-V1-7B  & 5.71&	5.57&	0.616 &	0.698 & 7.43 & 7.57 &	0.082 & 0.14 \\
    GPT-4o+UI-TARS-1.5-7B & 5.7	&7.53&	0.534 &	0.703 & 9.74 & 7.91&	0.169 & -1.83\\
    \rowcolor{gray!20} Avg. &2.72&3.51 &0.574 &0.672 & 6.02 & 5.23 &0.098 & -0.79  \\
    \bottomrule
  \end{tabularx}
  \caption{Additional metrics derived by SR and EQA. Here, $\text{EQ}^1_{15}$ and $\text{EQ}^2_{15}$ denotes for $\frac{EQA}{SR}$ and $\text{SR}-\text{EQA}$, respectively, when the maximal step is 15. $\Delta \text{EQ}^1 = \text{EQ}^1_{50} - \text{EQ}^1_{15}$ and so is the $\Delta \text{EQ}^2$. Similarly, $\Delta \text{SR} = \text{SR}_{50} - \text{SR}_{15}$ and $\Delta \text{EQA} = \text{EQA}_{50} - \text{EQA}_{15}$}
  \label{tab:insight3}
  \vspace{-3mm}
\end{table*}

We re-organize the $\frac{EQA}{SR}$ and $SR-EQA$ using the average results in Table~\ref{tab:singleapp-automation} as EQ$^1$ and EQ$^2$, and present the aggregated findings in Table~\ref{tab:insight3}. Combining with Figure~\ref{fig:EQA_visualization}, we can disclose four complementary patterns. First, the modular pairing of a powerful planner with a specialized grounder, exemplified by GPT-4o + UGround-V1-7B and GPT-4o + UI-TARS-1.5-7B, elevates the success rate under a 50-step budget by roughly 5.7\%, yet still incurs a substantial redundant step cost (EQ$^2$ = 7-8), signaling that cross-module coordination and early termination heuristics remain inadequate. Second, the large-scale DPO-aligned UI-TARS-72B-DPO achieves the strongest efficiency profile, increasing EQ$^1$ to 0.773 while compressing EQ$^2$ from 8.96 to 5.75 ($\Delta$EQ$^2$ = -3.21); this demonstrates that aligning to human preferences that explicitly reward rapid task completion can translate directly into tangible efficiency gains. Third, general-purpose agents such as GPT-4o and Claude-3.7 extract minimal benefit from a longer budget ($\Delta SR$<2.5\%) and even exhibit higher redundant step costs (EQ$^2$ increases from 3.09 to 3.22 and 3.40 to 3.65, respectively), underscoring that simply extending the interaction horizon cannot compensate for their limited visual granularity and action precision, therefore, integrating specialized perception or actuation modules is becoming indispensable. Lastly, none of the curves in Figure~\ref{fig:EQA_visualization} attains the ideal “hug-the-top-left-corner” profile, underscoring a pervasive lack of effective early-stopping heuristics and cost-aware search strategies.

To mitigate the efficiency bottlenecks aforementioned, we identify three possible research avenues. (1) Confidence- or value-based early-termination policies: equip agents with stopping rules that immediately end an episode when the marginal utility of further actions falls below a threshold, rather than passively consuming the entire step budget. (2) Cost-sensitive fine-tuning: during reinforcement-learning (or DPO-style) alignment, impose explicit penalties for every superfluous action so that optimization shifts from maximizing success rate (SR) alone to jointly maximizing the success-conditioned efficiency score EQA. (3) Progress-aware self-reflection: require the planner to periodically estimate the set of remaining sub-goals and, upon detecting that all objectives are satisfied, issue an immediate FINISH action. Together, these interventions target the twin goals of cutting redundant steps and encouraging agents to “know when to stop”, thereby narrowing the gap between current GUI agents and human-level operational efficiency.

\textbf{Finding 4: The limitation of action space restricts the agent’s ability to execute planned actions, especially in GUI task collaboration scenarios.} 

In Table~\ref{tab:multiapp-automation}, a notable fraction of models fail to complete the task on the web platform. The underlying cause is that, during web-interaction execution, the models lack the ability to trigger action $\texttt{switch\_tab}$ to enable 'press Tab to switch tabs'. In headless-browser settings, this omission blocks seamless navigation across multiple tabs, preventing cross-window information from being transferred from one context to another and ultimately derailing task completion.

On the other hand, due to the inherent heterogeneity of interactions across desktop, mobile, and web platforms, the current prompt-based definition of action functions struggles to comprehensively capture the full spectrum of platform-specific operations. Moreover, during inference, models may confuse actions across platforms, producing incorrect or incompatible output actions. 
Such issues can directly lead to task failure, even in single-platform, multi-app scenarios, and become particularly pronounced in multi-platform, multi-app settings, for example, when copying text from a web page and pasting it into a desktop application like Word for further formatting.

Building on these observations, we argue that a more generalizable, extensible, and potentially platform-agnostic definition of the action space is worth pursuing.
One intuitive and straightforward direction is to construct a unified API abstraction layer that comprehensively covers multi-platform operations. Under this design, the agent interacts with the environment by invoking platform-independent APIs, while the backend of the API is responsible for platform-specific adaptations.
An alternative route focuses on operation atomization. Unlike current action spaces that rely on fixed, platform-tied commands, an ideal action space would emphasize a set of primitive operations, decoupled from any particular environment. Agent-issued instructions are then mapped to these primitives via a many-to-many translation schema, where each high-level intent may correspond to a combination of atomic steps. These atomic units can then be recompiled into platform-specific execution commands, enabling robust and consistent interaction across environments.
Beyond these two approaches, we believe that the research community should continue to explore better formulations of the action space, those characterized by strong generality, high extensibility, and minimal platform dependence.

\textbf{Finding 5: Although many GUI agents excel in simple cases, their effectiveness diminishes significantly as task complexity rises, revealing limited generalization capabilities.
}

\begin{figure*}[!t]
    \centering
    \includegraphics[width=\linewidth]{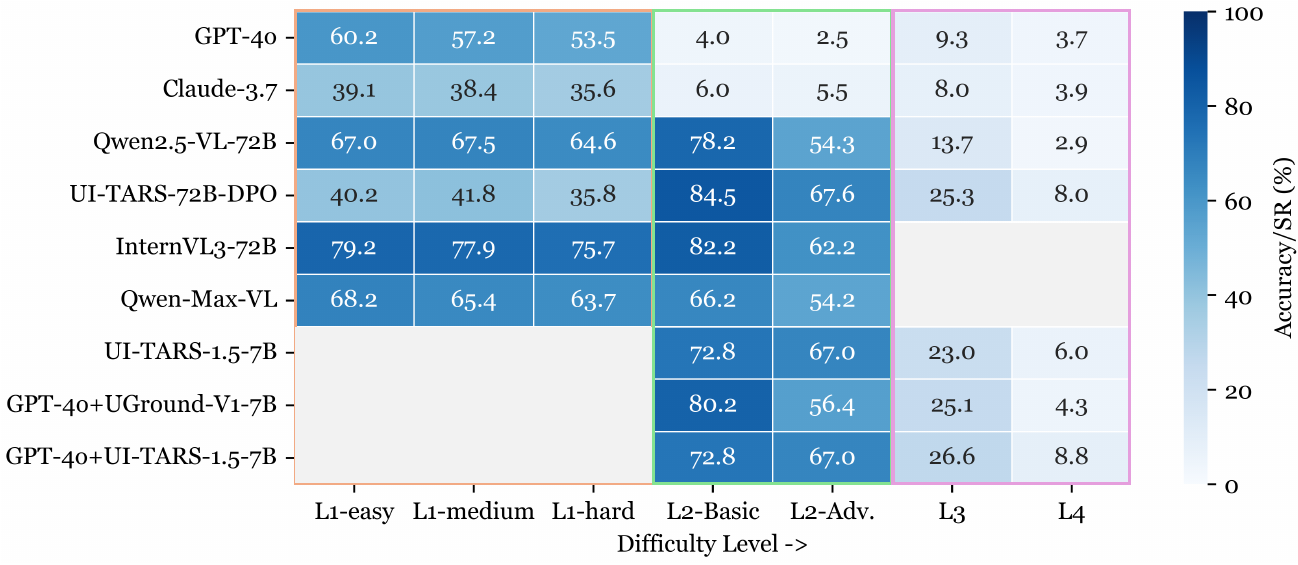}
    \caption{\textbf{Difficulty-Gradient Heatmap.} Models’ scores across difficulty levels are encoded with a single-hue palette whose saturation fades from high (dark) to low (light). Colored rectangles outline comparable model groups. Within and across these groups, the color consistently fades from L1, L2 to L3 and L4, indicating that higher task complexity amplifies each model’s weaknesses and causes a steep performance drop-off.}
    \vspace{-2mm}
    \label{fig: Difficulty_Gradient_Heatmap}
\end{figure*}

As shown in Figure~\ref{fig: Difficulty_Gradient_Heatmap}, although many systems perform impressively on easy scenarios, their accuracy/success rate deteriorates sharply as soon as either (i) the local difficulty within a level increases (easy $\rightarrow$ medium $\rightarrow$ hard; basic $\rightarrow$ advanced) or (ii) the global task complexity rises from L1 to L4.  These steep drops - especially pronounced for general-purpose LLMs - indicate that today’s agents still lack robust generalization to harder, less stereotyped GUI situations. For example, the GUI understanding score of GPT-4o drops from 60.2\% (easy) to 53.5\% (hard), a -11\% decrease, while even the highly tuned InternVL3-72B loses 4\% (Table \ref{tab:understanding}). In element grounding, switching from 'Basic' to semantically implicit 'Advanced' queries slashes GPT-4o’s mean accuracy by nearly 40\% and still costs the specialist UI-TARS-72B-DPO 16\% (Table \ref{tab:grounding}).  The effect compounds across levels: the strongest agent (GPT-4o + UI-TARS-1.5-7B) succeeds in 26.6\% of tasks at L3 but only 8.8\% once multi-app collaboration is required in L4, a 67\% collapse that is mirrored by other models (Tables \ref{tab:singleapp-automation}–\ref{tab:multiapp-automation}). Concomitant declines in EQA confirm that agents not only fail more often but also waste proportionally more steps before failing.

These sharp drops expose three intertwined bottlenecks: (1) ill-posed perceptual clues (small widgets, non-salient text), (2) longer credit-assignment chains, and (3) noisy action spaces inflate the search space exponentially. Current models, trained largely on static screenshots, lack the robust abstract representations and error-driven exploration strategies needed to cope. 

Possible targeted remedies include: (1) Curriculum \& hard-negative mining.  Intentionally up-sample adversarial layouts (occlusion, theme changes, deceptive affordances) during instruction tuning to inoculate perceptual modules against distribution shift. (2) Dynamic skill routing. Teach planners to self-diagnose uncertainty and automatically invoke auxiliary skills (OCR, vision transformers, memory retrieval) as difficulty rises. (3) Hierarchical planners with macro-actions.  Introduce option-level abstractions (e.g., \texttt{open-browser-tab}) so that sparse EQA-style rewards can flow to high-level decisions instead of individual clicks. (4) Unified state schema for all applications.  Store “App → Page → Element” graphs in an external memory that survives context switches, allowing the planner to reason over shared entities rather than raw pixel buffers.

We believe that by attacking these verified failure modes, the community can turn today’s hardest cases, from implicitly described buttons to multi-window workflows, into stepping-stones toward truly general-purpose GUI agents.

\textbf{Finding 6: The failures in multi-application environments primarily stem from limited cross-context memory and action space, rather than issues with perception or planning.}

Success drops that cannot be explained by harder screenshots or longer action chains alone appear as soon as the agent must pass information between applications. The strongest single-app system, GPT-4o+UI-TARS-1.5-7B, falls from \textbf{26.6\%} SR on L3 to just \textbf{8.8\%} on L4 (Tables \ref{tab:singleapp-automation}–\ref{tab:multiapp-automation}); UI-TARS-72B-DPO shows an almost identical collapse (25.3\% to 8.0\%). Failures concentrate at window or tab boundaries: five models are labeled `-' on the Web platform simply because they cannot express the primitive \texttt{switch\_tab}. At the same time, EQA shrinks far more than the accompanying $SR-EQA$ penalty (e.g., 18.7\% → 6.4\% for GPT-4o + UI-TARS), signaling that agents waste many steps rediscovering the context they have just lost. These phenomena point to a deficit in working memory and action-space coverage, rather than in perception or generic planning.

Addressing these failures may require agents to focus on memory-centric research avenues, including: (1) External episodic buffer. Log every UI observation and write-back (\textit{copy}, \textit{navigate}, \textit{paste} …) to an append-only timeline that the language planner can query with natural language—much like retrieval-augmented generation, but for GUI states. (2) Semantic anchors.  Tag entities (e.g., “flight-price \$514”) with stable IDs when first seen; subsequent references use the anchor, so the planner no longer depends on window focus to recall an object. (3) Cross-context consistency checks. Inject lightweight assertions, for example, “clipboard should now contain X” and “target window title equals Y”. Violations trigger immediate self-repair instead of long, fruitless trial-and-error loops, cutting the redundant steps that dominate L4 failures.

\section{Conclusion}

In this work, we presented MMBench-GUI, a novel hierarchical multi-platform evaluation framework that comprehensively assesses the capabilities and limitations of GUI automation agents. Through rigorous evaluations across multiple operating systems and diverse tasks, we uncovered critical insights into key performance bottlenecks, particularly highlighting the importance of accurate visual grounding, sophisticated planning, and robust cross-platform generalization. Our findings demonstrate that modular architectures integrating specialized grounding modules significantly improve performance, addressing inherent limitations of general-purpose language models. Additionally, our analysis underscores the importance of improving long-horizon reasoning, adaptive error recovery, and effective memory and state management to address complex and ambiguous GUI scenarios. MMBench-GUI thus provides a foundational benchmarking resource and actionable guidance for future research efforts, advancing the development of robust, reliable, and practically applicable GUI automation agents.

\subsubsection*{Future work}
We will strengthen our study along three aspects:
(1) Broader model coverage. We will evaluate a wider spectrum of models—including open-source, proprietary, and the latest RL-based systems—so that each model is tested across all difficulty levels. (2) Deeper analysis. With a richer experimental pool, we will perform fine-grained analyses to produce more robust and generalizable findings. (3) Task expansion \& error attribution. We plan to add more online tasks to cover a broader set of applications, validate their correctness step by step, and log sufficient runtime details to pinpoint the exact causes of failure.

\clearpage

\bibliography{iclr2025_conference}

\begin{thebibliography}{55}
\providecommand{\natexlab}[1]{#1}
\providecommand{\url}[1]{\texttt{#1}}
\expandafter\ifx\csname urlstyle\endcsname\relax
  \providecommand{\doi}[1]{doi: #1}\else
  \providecommand{\doi}{doi: \begingroup \urlstyle{rm}\Url}\fi

\bibitem[Anthropic(2024)]{claude35}
Sonnet Anthropic.
\newblock Model card addendum: Claude 3.5 haiku and upgraded claude 3.5 sonnet.
\newblock 2024.
\newblock URL \url{https://api.semanticscholar.org/CorpusID:273639283}.

\bibitem[Anthropic(2025)]{anthropic2025claude37}
Sonnet Anthropic.
\newblock Claude 3.7 sonnet system card.
\newblock 2025.
\newblock URL \url{https://www.anthropic.com/news/claude-3-7-sonnet}.

\bibitem[Bai et~al.(2023)Bai, Bai, Yang, Wang, Tan, Wang, Lin, Zhou, and Zhou]{Qwen-VL}
Jinze Bai, Shuai Bai, Shusheng Yang, Shijie Wang, Sinan Tan, Peng Wang, Junyang Lin, Chang Zhou, and Jingren Zhou.
\newblock Qwen-vl: A frontier large vision-language model with versatile abilities.
\newblock \emph{arXiv preprint arXiv:2308.12966}, 2023.

\bibitem[Bai et~al.(2025)Bai, Chen, Liu, Wang, Ge, Song, Dang, Wang, Wang, Tang, et~al.]{bai2025qwen25vl}
Shuai Bai, Keqin Chen, Xuejing Liu, Jialin Wang, Wenbin Ge, Sibo Song, Kai Dang, Peng Wang, Shijie Wang, Jun Tang, et~al.
\newblock Qwen2. 5-vl technical report.
\newblock \emph{arXiv preprint arXiv:2502.13923}, 2025.

\bibitem[Bonatti et~al.(2024)Bonatti, Zhao, Bonacci, Dupont, Abdali, Li, Lu, Wagle, Koishida, Bucker, et~al.]{bonatti2024windows}
Rogerio Bonatti, Dan Zhao, Francesco Bonacci, Dillon Dupont, Sara Abdali, Yinheng Li, Yadong Lu, Justin Wagle, Kazuhito Koishida, Arthur Bucker, et~al.
\newblock Windows agent arena: Evaluating multi-modal os agents at scale.
\newblock \emph{arXiv preprint arXiv:2409.08264}, 2024.

\bibitem[Chang et~al.(2024)Chang, Zhang, Zhu, Yang, Yang, Jin, Lan, Kong, and He]{chang2024agentboard}
Ma~Chang, Junlei Zhang, Zhihao Zhu, Cheng Yang, Yujiu Yang, Yaohui Jin, Zhenzhong Lan, Lingpeng Kong, and Junxian He.
\newblock Agentboard: An analytical evaluation board of multi-turn llm agents.
\newblock \emph{Advances in neural information processing systems}, 37:\penalty0 74325--74362, 2024.

\bibitem[Chen et~al.(2024{\natexlab{a}})Chen, Cui, Hu, Qin, Fang, Zhao, Wang, Liu, Chen, Huo, et~al.]{chen2024guicourse}
Wentong Chen, Junbo Cui, Jinyi Hu, Yujia Qin, Junjie Fang, Yue Zhao, Chongyi Wang, Jun Liu, Guirong Chen, Yupeng Huo, et~al.
\newblock Guicourse: From general vision language models to versatile gui agents.
\newblock \emph{arXiv preprint arXiv:2406.11317}, 2024{\natexlab{a}}.

\bibitem[Chen et~al.(2021)Chen, Zhao, Chen, Zhang, Ji, Luo, Xiong, and Yu]{chen2021websrc}
Xingyu Chen, Zihan Zhao, Lu~Chen, Danyang Zhang, Jiabao Ji, Ao~Luo, Yuxuan Xiong, and Kai Yu.
\newblock Websrc: a dataset for web-based structural reading comprehension.
\newblock \emph{arXiv preprint arXiv:2101.09465}, 2021.

\bibitem[Chen et~al.(2024{\natexlab{b}})Chen, Wang, Cao, Liu, Gao, Cui, Zhu, Ye, Tian, Liu, et~al.]{chen2024expanding}
Zhe Chen, Weiyun Wang, Yue Cao, Yangzhou Liu, Zhangwei Gao, Erfei Cui, Jinguo Zhu, Shenglong Ye, Hao Tian, Zhaoyang Liu, et~al.
\newblock Expanding performance boundaries of open-source multimodal models with model, data, and test-time scaling.
\newblock \emph{arXiv preprint arXiv:2412.05271}, 2024{\natexlab{b}}.

\bibitem[Cheng et~al.(2024)Cheng, Sun, Chu, Xu, YanTao, Zhang, and Wu]{cheng2024seeclick}
Kanzhi Cheng, Qiushi Sun, Yougang Chu, Fangzhi Xu, Li~YanTao, Jianbing Zhang, and Zhiyong Wu.
\newblock {S}ee{C}lick: Harnessing {GUI} grounding for advanced visual {GUI} agents.
\newblock In \emph{Proceedings of the 62nd Annual Meeting of the Association for Computational Linguistics (Volume 1: Long Papers)}, pp.\  9313--9332, Bangkok, Thailand, August 2024. Association for Computational Linguistics.
\newblock \doi{10.18653/v1/2024.acl-long.505}.
\newblock URL \url{https://aclanthology.org/2024.acl-long.505/}.

\bibitem[Deng et~al.(2023{\natexlab{a}})Deng, Gu, Zheng, Chen, Stevens, Wang, Sun, and Su]{deng2023mind2web}
Xiang Deng, Yu~Gu, Boyuan Zheng, Shijie Chen, Sam Stevens, Boshi Wang, Huan Sun, and Yu~Su.
\newblock Mind2web: Towards a generalist agent for the web.
\newblock \emph{Advances in Neural Information Processing Systems}, 36:\penalty0 28091--28114, 2023{\natexlab{a}}.

\bibitem[Deng et~al.(2023{\natexlab{b}})Deng, Gu, Zheng, Chen, Stevens, Wang, Sun, and Su]{deng2023mindweb}
Xiang Deng, Yu~Gu, Boyuan Zheng, Shijie Chen, Samuel Stevens, Boshi Wang, Huan Sun, and Yu~Su.
\newblock Mind2web: Towards a generalist agent for the web.
\newblock In \emph{Thirty-seventh Conference on Neural Information Processing Systems Datasets and Benchmarks Track}, 2023{\natexlab{b}}.
\newblock URL \url{https://openreview.net/forum?id=kiYqbO3wqw}.

\bibitem[Furuta et~al.(2023)Furuta, Lee, Nachum, Matsuo, Faust, Gu, and Gur]{furuta2023multimodal}
Hiroki Furuta, Kuang-Huei Lee, Ofir Nachum, Yutaka Matsuo, Aleksandra Faust, Shixiang~Shane Gu, and Izzeddin Gur.
\newblock Multimodal web navigation with instruction-finetuned foundation models.
\newblock \emph{arXiv preprint arXiv:2305.11854}, 2023.

\bibitem[Gao et~al.(2023)Gao, Ji, Bai, Ouyang, Li, Mao, Wu, Zhang, Wang, Guo, et~al.]{gao2023assistgui}
Difei Gao, Lei Ji, Zechen Bai, Mingyu Ouyang, Peiran Li, Dongxing Mao, Qinchen Wu, Weichen Zhang, Peiyi Wang, Xiangwu Guo, et~al.
\newblock Assistgui: Task-oriented desktop graphical user interface automation.
\newblock \emph{arXiv preprint arXiv:2312.13108}, 2023.

\bibitem[Gou et~al.(2024)Gou, Wang, Zheng, Xie, Chang, Shu, Sun, and Su]{gou2024navigating}
Boyu Gou, Ruohan Wang, Boyuan Zheng, Yanan Xie, Cheng Chang, Yiheng Shu, Huan Sun, and Yu~Su.
\newblock Navigating the digital world as humans do: Universal visual grounding for gui agents.
\newblock \emph{arXiv preprint arXiv:2410.05243}, 2024.

\bibitem[He et~al.(2024)He, Yao, Ma, Yu, Dai, Zhang, Lan, and Yu]{he2024webvoyager}
Hongliang He, Wenlin Yao, Kaixin Ma, Wenhao Yu, Yong Dai, Hongming Zhang, Zhenzhong Lan, and Dong Yu.
\newblock Webvoyager: Building an end-to-end web agent with large multimodal models.
\newblock \emph{arXiv preprint arXiv:2401.13919}, 2024.

\bibitem[Hong et~al.(2024)Hong, Wang, Lv, Xu, Yu, Ji, Wang, Wang, Dong, Ding, et~al.]{hong2024cogagent}
Wenyi Hong, Weihan Wang, Qingsong Lv, Jiazheng Xu, Wenmeng Yu, Junhui Ji, Yan Wang, Zihan Wang, Yuxiao Dong, Ming Ding, et~al.
\newblock Cogagent: A visual language model for gui agents.
\newblock In \emph{Proceedings of the IEEE/CVF Conference on Computer Vision and Pattern Recognition}, pp.\  14281--14290, 2024.

\bibitem[Hsiao et~al.(2022)Hsiao, Zubach, Baechler, Carbune, Lin, Wang, Sunkara, Zhu, and Chen]{hsiao2022screenqa}
Yu-Chung Hsiao, Fedir Zubach, Gilles Baechler, Victor Carbune, Jason Lin, Maria Wang, Srinivas Sunkara, Yun Zhu, and Jindong Chen.
\newblock Screenqa: Large-scale question-answer pairs over mobile app screenshots.
\newblock \emph{arXiv preprint arXiv:2209.08199}, 2022.

\bibitem[Hurst et~al.(2024)Hurst, Lerer, Goucher, Perelman, Ramesh, Clark, Ostrow, Welihinda, Hayes, Radford, et~al.]{hurst2024gpt}
Aaron Hurst, Adam Lerer, Adam~P Goucher, Adam Perelman, Aditya Ramesh, Aidan Clark, AJ~Ostrow, Akila Welihinda, Alan Hayes, Alec Radford, et~al.
\newblock Gpt-4o system card.
\newblock \emph{arXiv preprint arXiv:2410.21276}, 2024.

\bibitem[Kapoor et~al.(2024)Kapoor, Butala, Russak, Koh, Kamble, AlShikh, and Salakhutdinov]{kapoor2024omniact}
Raghav Kapoor, Yash~Parag Butala, Melisa Russak, Jing~Yu Koh, Kiran Kamble, Waseem AlShikh, and Ruslan Salakhutdinov.
\newblock Omniact: A dataset and benchmark for enabling multimodal generalist autonomous agents for desktop and web.
\newblock In \emph{European Conference on Computer Vision}, pp.\  161--178. Springer, 2024.

\bibitem[Li et~al.(2025)Li, Meng, Lin, Luo, Tian, Ma, Huang, and Chua]{li2025screenspotpro}
Kaixin Li, Ziyang Meng, Hongzhan Lin, Ziyang Luo, Yuchen Tian, Jing Ma, Zhiyong Huang, and Tat-Seng Chua.
\newblock Screenspot-pro: Gui grounding for professional high-resolution computer use.
\newblock \emph{arXiv preprint arXiv:2504.07981}, 2025.

\bibitem[Li et~al.(2024)Li, Bishop, Li, Rawles, Campbell-Ajala, Tyamagundlu, and Riva]{li2024effects}
Wei Li, William~E Bishop, Alice Li, Christopher Rawles, Folawiyo Campbell-Ajala, Divya Tyamagundlu, and Oriana Riva.
\newblock On the effects of data scale on ui control agents.
\newblock \emph{Advances in Neural Information Processing Systems}, 37:\penalty0 92130--92154, 2024.

\bibitem[Lin et~al.(2024)Lin, Li, Gao, Yang, Wu, Bai, Lei, Wang, and Shou]{lin2024showui}
Kevin~Qinghong Lin, Linjie Li, Difei Gao, Zhengyuan Yang, Shiwei Wu, Zechen Bai, Weixian Lei, Lijuan Wang, and Mike~Zheng Shou.
\newblock Showui: One vision-language-action model for gui visual agent, 2024.
\newblock URL \url{https://arxiv.org/abs/2411.17465}.

\bibitem[Lin et~al.(2014)Lin, Maire, Belongie, Hays, Perona, Ramanan, Doll{\'a}r, and Zitnick]{lin2014microsoft}
Tsung-Yi Lin, Michael Maire, Serge Belongie, James Hays, Pietro Perona, Deva Ramanan, Piotr Doll{\'a}r, and C~Lawrence Zitnick.
\newblock Microsoft coco: Common objects in context.
\newblock In \emph{European conference on computer vision}, pp.\  740--755. Springer, 2014.

\bibitem[Liu et~al.(2024{\natexlab{a}})Liu, Qin, Liang, Dong, Lai, Zhang, Zhao, Iong, Sun, Wang, et~al.]{liu2024autoglm}
Xiao Liu, Bo~Qin, Dongzhu Liang, Guang Dong, Hanyu Lai, Hanchen Zhang, Hanlin Zhao, Iat~Long Iong, Jiadai Sun, Jiaqi Wang, et~al.
\newblock Autoglm: Autonomous foundation agents for guis.
\newblock \emph{arXiv preprint arXiv:2411.00820}, 2024{\natexlab{a}}.

\bibitem[Liu et~al.(2024{\natexlab{b}})Liu, Zhang, Gu, Iong, Xu, Song, Zhang, Lai, Liu, Zhao, et~al.]{liu2024visualagentbench}
Xiao Liu, Tianjie Zhang, Yu~Gu, Iat~Long Iong, Yifan Xu, Xixuan Song, Shudan Zhang, Hanyu Lai, Xinyi Liu, Hanlin Zhao, et~al.
\newblock Visualagentbench: Towards large multimodal models as visual foundation agents.
\newblock \emph{arXiv preprint arXiv:2408.06327}, 2024{\natexlab{b}}.

\bibitem[Liu et~al.(2025)Liu, Zhang, Zhang, and Lu]{liu2025ui}
Xinyi Liu, Xiaoyi Zhang, Ziyun Zhang, and Yan Lu.
\newblock Ui-e2i-synth: Advancing gui grounding with large-scale instruction synthesis.
\newblock \emph{arXiv preprint arXiv:2504.11257}, 2025.

\bibitem[Liu et~al.(2024{\natexlab{c}})Liu, Duan, Zhang, Li, Zhang, Zhao, Yuan, Wang, He, Liu, et~al.]{liu2024mmbench}
Yuan Liu, Haodong Duan, Yuanhan Zhang, Bo~Li, Songyang Zhang, Wangbo Zhao, Yike Yuan, Jiaqi Wang, Conghui He, Ziwei Liu, et~al.
\newblock Mmbench: Is your multi-modal model an all-around player?
\newblock In \emph{European conference on computer vision}, pp.\  216--233. Springer, 2024{\natexlab{c}}.

\bibitem[Lu et~al.(2024)Lu, Shao, Liu, Meng, Li, Chen, Huang, Zhang, Qiao, and Luo]{lu2024guiodyssey}
Quanfeng Lu, Wenqi Shao, Zitao Liu, Fanqing Meng, Boxuan Li, Botong Chen, Siyuan Huang, Kaipeng Zhang, Yu~Qiao, and Ping Luo.
\newblock Gui odyssey: A comprehensive dataset for cross-app gui navigation on mobile devices.
\newblock \emph{arXiv preprint arXiv:2406.08451}, 2024.

\bibitem[Masry et~al.(2022)Masry, Long, Tan, Joty, and Hoque]{masry2022chartqa}
Ahmed Masry, Do~Xuan Long, Jia~Qing Tan, Shafiq Joty, and Enamul Hoque.
\newblock Chartqa: A benchmark for question answering about charts with visual and logical reasoning.
\newblock \emph{arXiv preprint arXiv:2203.10244}, 2022.

\bibitem[Nayak et~al.(2025)Nayak, Jian, Lin, Rodriguez, Kalsi, Awal, Chapados, {\"O}zsu, Agrawal, Vazquez, et~al.]{nayak2025ui}
Shravan Nayak, Xiangru Jian, Kevin~Qinghong Lin, Juan~A Rodriguez, Montek Kalsi, Rabiul Awal, Nicolas Chapados, M~Tamer {\"O}zsu, Aishwarya Agrawal, David Vazquez, et~al.
\newblock Ui-vision: A desktop-centric gui benchmark for visual perception and interaction.
\newblock \emph{arXiv preprint arXiv:2503.15661}, 2025.

\bibitem[Niu et~al.(2024)Niu, Li, Wang, Fu, Hu, Leng, Kong, Chang, and Wang]{niu2024screenagent}
Runliang Niu, Jindong Li, Shiqi Wang, Yali Fu, Xiyu Hu, Xueyuan Leng, He~Kong, Yi~Chang, and Qi~Wang.
\newblock Screenagent: A vision language model-driven computer control agent.
\newblock 2024.

\bibitem[OpenAI(2025)]{openai2025introducing}
OpenAI.
\newblock Introducing openai o3 and o4-mini.
\newblock \url{https://openai.com/index/introducing-o3-and-o4-mini}, 2025.

\bibitem[Qin et~al.(2025)Qin, Ye, Fang, Wang, Liang, Tian, Zhang, Li, Li, Huang, et~al.]{qin2025ui}
Yujia Qin, Yining Ye, Junjie Fang, Haoming Wang, Shihao Liang, Shizuo Tian, Junda Zhang, Jiahao Li, Yunxin Li, Shijue Huang, et~al.
\newblock Ui-tars: Pioneering automated gui interaction with native agents.
\newblock \emph{arXiv preprint arXiv:2501.12326}, 2025.

\bibitem[Rawles et~al.(2023)Rawles, Li, Rodriguez, Riva, and Lillicrap]{rawles2023androidinthewild}
Christopher Rawles, Alice Li, Daniel Rodriguez, Oriana Riva, and Timothy Lillicrap.
\newblock Androidinthewild: A large-scale dataset for android device control.
\newblock \emph{Advances in Neural Information Processing Systems}, 36:\penalty0 59708--59728, 2023.

\bibitem[Rawles et~al.(2024)Rawles, Clinckemaillie, Chang, Waltz, Lau, Fair, Li, Bishop, Li, Campbell-Ajala, et~al.]{rawles2024androidworld}
Christopher Rawles, Sarah Clinckemaillie, Yifan Chang, Jonathan Waltz, Gabrielle Lau, Marybeth Fair, Alice Li, William Bishop, Wei Li, Folawiyo Campbell-Ajala, et~al.
\newblock Androidworld: A dynamic benchmarking environment for autonomous agents.
\newblock \emph{arXiv preprint arXiv:2405.14573}, 2024.

\bibitem[Sun et~al.(2024)Sun, Cheng, Ding, Jin, Wang, Xu, Wu, Jia, Chen, Liu, et~al.]{sun2024osgenesis}
Qiushi Sun, Kanzhi Cheng, Zichen Ding, Chuanyang Jin, Yian Wang, Fangzhi Xu, Zhenyu Wu, Chengyou Jia, Liheng Chen, Zhoumianze Liu, et~al.
\newblock Os-genesis: Automating gui agent trajectory construction via reverse task synthesis.
\newblock \emph{arXiv preprint arXiv:2412.19723}, 2024.

\bibitem[Sun et~al.(2025)Sun, Liu, Ma, Ding, Xu, Yin, Zhao, Wu, Cheng, Liu, et~al.]{sun2025scienceboard}
Qiushi Sun, Zhoumianze Liu, Chang Ma, Zichen Ding, Fangzhi Xu, Zhangyue Yin, Haiteng Zhao, Zhenyu Wu, Kanzhi Cheng, Zhaoyang Liu, et~al.
\newblock Scienceboard: Evaluating multimodal autonomous agents in realistic scientific workflows.
\newblock \emph{arXiv preprint arXiv:2505.19897}, 2025.

\bibitem[Team et~al.(2025)Team, Du, Yin, Xing, Qu, Wang, Chen, Zhang, Du, Wei, et~al.]{team2025kimi}
Kimi Team, Angang Du, Bohong Yin, Bowei Xing, Bowen Qu, Bowen Wang, Cheng Chen, Chenlin Zhang, Chenzhuang Du, Chu Wei, et~al.
\newblock Kimi-vl technical report.
\newblock \emph{arXiv preprint arXiv:2504.07491}, 2025.

\bibitem[Wang et~al.(2025)Wang, Wang, Deng, Xie, Li, Zhang, Li, Hua, Su, Yang, Zhang, Wang, Zhong, and Yu]{wang2025computer}
Bowen Wang, Xinyuan Wang, Jiaqi Deng, Tianbao Xie, Ryan Li, Yanzhe Zhang, Gavin Li, Toh~Jing Hua, Yu~Su, Diyi Yang, Yi~Zhang, Zhiguo Wang, Victor Zhong, and Tao Yu.
\newblock Computer agent arena: Compare \& test computer use agents on crowdsourced real-world tasks, 2025.
\newblock URL \url{https://arena.xlang.ai}.

\bibitem[Wang et~al.(2024)Wang, Bai, Tan, Wang, Fan, Bai, Chen, Liu, Wang, Ge, et~al.]{wang2024qwen2vl}
Peng Wang, Shuai Bai, Sinan Tan, Shijie Wang, Zhihao Fan, Jinze Bai, Keqin Chen, Xuejing Liu, Jialin Wang, Wenbin Ge, et~al.
\newblock Qwen2-vl: Enhancing vision-language model's perception of the world at any resolution.
\newblock \emph{arXiv preprint arXiv:2409.12191}, 2024.

\bibitem[Wu et~al.(2025)Wu, Cheng, Yang, Zhang, Yang, Jiang, Mu, Peng, Qiao, Tan, et~al.]{wu2025gui}
Qianhui Wu, Kanzhi Cheng, Rui Yang, Chaoyun Zhang, Jianwei Yang, Huiqiang Jiang, Jian Mu, Baolin Peng, Bo~Qiao, Reuben Tan, et~al.
\newblock Gui-actor: Coordinate-free visual grounding for gui agents.
\newblock \emph{arXiv preprint arXiv:2506.03143}, 2025.

\bibitem[Wu et~al.(2024{\natexlab{a}})Wu, Han, Ding, Weng, Liu, Yao, Yu, and Kong]{wu2024copilot}
Zhiyong Wu, Chengcheng Han, Zichen Ding, Zhenmin Weng, Zhoumianze Liu, Shunyu Yao, Tao Yu, and Lingpeng Kong.
\newblock Os-copilot: Towards generalist computer agents with self-improvement.
\newblock \emph{arXiv preprint arXiv:2402.07456}, 2024{\natexlab{a}}.

\bibitem[Wu et~al.(2024{\natexlab{b}})Wu, Wu, Xu, Wang, Sun, Jia, Cheng, Ding, Chen, Liang, et~al.]{wu2024atlas}
Zhiyong Wu, Zhenyu Wu, Fangzhi Xu, Yian Wang, Qiushi Sun, Chengyou Jia, Kanzhi Cheng, Zichen Ding, Liheng Chen, Paul~Pu Liang, et~al.
\newblock Os-atlas: A foundation action model for generalist gui agents.
\newblock \emph{arXiv preprint arXiv:2410.23218}, 2024{\natexlab{b}}.

\bibitem[Xiaomi(2025)]{coreteam2025mimovltechnicalreport}
LLM-Core-Team Xiaomi.
\newblock Mimo-vl technical report, 2025.
\newblock URL \url{https://arxiv.org/abs/2506.03569}.

\bibitem[Xie et~al.(2024)Xie, Zhang, Chen, Li, Zhao, Cao, Hua, Cheng, Shin, Lei, et~al.]{xie2024osworld}
Tianbao Xie, Danyang Zhang, Jixuan Chen, Xiaochuan Li, Siheng Zhao, Ruisheng Cao, Toh~J Hua, Zhoujun Cheng, Dongchan Shin, Fangyu Lei, et~al.
\newblock Osworld: Benchmarking multimodal agents for open-ended tasks in real computer environments.
\newblock \emph{Advances in Neural Information Processing Systems}, 37:\penalty0 52040--52094, 2024.

\bibitem[Xie et~al.(2025)Xie, Deng, Li, Yang, Wu, Chen, Hu, Wang, Xu, Wang, et~al.]{xie2025scaling}
Tianbao Xie, Jiaqi Deng, Xiaochuan Li, Junlin Yang, Haoyuan Wu, Jixuan Chen, Wenjing Hu, Xinyuan Wang, Yuhui Xu, Zekun Wang, et~al.
\newblock Scaling computer-use grounding via user interface decomposition and synthesis.
\newblock \emph{arXiv preprint arXiv:2505.13227}, 2025.

\bibitem[Xu et~al.(2024{\natexlab{a}})Xu, Liu, Sun, Cheng, Yu, Lai, Zhang, Zhang, Tang, and Dong]{xu2024androidlab}
Yifan Xu, Xiao Liu, Xueqiao Sun, Siyi Cheng, Hao Yu, Hanyu Lai, Shudan Zhang, Dan Zhang, Jie Tang, and Yuxiao Dong.
\newblock Androidlab: Training and systematic benchmarking of android autonomous agents.
\newblock \emph{arXiv preprint arXiv:2410.24024}, 2024{\natexlab{a}}.

\bibitem[Xu et~al.(2024{\natexlab{b}})Xu, Wang, Wang, Lu, Xie, Saha, Sahoo, Yu, and Xiong]{xu2024aguvis}
Yiheng Xu, Zekun Wang, Junli Wang, Dunjie Lu, Tianbao Xie, Amrita Saha, Doyen Sahoo, Tao Yu, and Caiming Xiong.
\newblock Aguvis: Unified pure vision agents for autonomous gui interaction.
\newblock \emph{arXiv preprint arXiv:2412.04454}, 2024{\natexlab{b}}.

\bibitem[Yang et~al.(2024)Yang, Wang, Li, Luo, Chen, Huang, and Li]{yang2024aria}
Yuhao Yang, Yue Wang, Dongxu Li, Ziyang Luo, Bei Chen, Chao Huang, and Junnan Li.
\newblock Aria-ui: Visual grounding for gui instructions.
\newblock \emph{arXiv preprint arXiv:2412.16256}, 2024.

\bibitem[Yue et~al.(2024)Yue, Ni, Zhang, Zheng, Liu, Zhang, Stevens, Jiang, Ren, Sun, et~al.]{yue2024mmmu}
Xiang Yue, Yuansheng Ni, Kai Zhang, Tianyu Zheng, Ruoqi Liu, Ge~Zhang, Samuel Stevens, Dongfu Jiang, Weiming Ren, Yuxuan Sun, et~al.
\newblock Mmmu: A massive multi-discipline multimodal understanding and reasoning benchmark for expert agi.
\newblock In \emph{Proceedings of the IEEE/CVF Conference on Computer Vision and Pattern Recognition}, pp.\  9556--9567, 2024.

\bibitem[Zhang et~al.(2025)Zhang, Ding, Ma, Chen, Sun, Lan, and He]{zhang2025guimid}
Junlei Zhang, Zichen Ding, Chang Ma, Zijie Chen, Qiushi Sun, Zhenzhong Lan, and Junxian He.
\newblock Breaking the data barrier--building gui agents through task generalization.
\newblock \emph{arXiv preprint arXiv:2504.10127}, 2025.

\bibitem[Zheng et~al.(2024)Zheng, Gou, Kil, Sun, and Su]{zheng2024seeact}
Boyuan Zheng, Boyu Gou, Jihyung Kil, Huan Sun, and Yu~Su.
\newblock Gpt-4v(ision) is a generalist web agent, if grounded.
\newblock In \emph{Forty-first International Conference on Machine Learning}, 2024.
\newblock URL \url{https://openreview.net/forum?id=piecKJ2DlB}.

\bibitem[Zhou et~al.(2023)Zhou, Xu, Zhu, Zhou, Lo, Sridhar, Cheng, Ou, Bisk, Fried, et~al.]{zhou2023webarena}
Shuyan Zhou, Frank~F Xu, Hao Zhu, Xuhui Zhou, Robert Lo, Abishek Sridhar, Xianyi Cheng, Tianyue Ou, Yonatan Bisk, Daniel Fried, et~al.
\newblock Webarena: A realistic web environment for building autonomous agents.
\newblock \emph{arXiv preprint arXiv:2307.13854}, 2023.

\bibitem[Zhu et~al.(2025)Zhu, Wang, Chen, Liu, Ye, Gu, Duan, Tian, Su, Shao, et~al.]{zhu2025internvl3}
Jinguo Zhu, Weiyun Wang, Zhe Chen, Zhaoyang Liu, Shenglong Ye, Lixin Gu, Yuchen Duan, Hao Tian, Weijie Su, Jie Shao, et~al.
\newblock Internvl3: Exploring advanced training and test-time recipes for open-source multimodal models.
\newblock \emph{arXiv preprint arXiv:2504.10479}, 2025.

\end{thebibliography}
\bibliographystyle{iclr2025_conference}

\appendix


\end{document}